\documentclass{article} 
\usepackage{iclr2027_conference,times}

\usepackage{amsmath,amsfonts,bm}

\def\eqref#1{equation~\ref{#1}}

\def\1{\bm{1}}

\DeclareMathAlphabet{\mathsfit}{\encodingdefault}{\sfdefault}{m}{sl}
\SetMathAlphabet{\mathsfit}{bold}{\encodingdefault}{\sfdefault}{bx}{n}

\usepackage{algorithm}
\usepackage{algpseudocode}
\usepackage{graphicx}
\usepackage{booktabs}
\usepackage{multirow}
\usepackage{placeins}
\numberwithin{equation}{section}
\renewcommand{\theequation}{\arabic{equation}}
\numberwithin{algorithm}{section}

\AtBeginDocument{%
  \raggedbottom
  \setlength{\abovedisplayskip}{4pt}%
  \setlength{\belowdisplayskip}{4pt}%
  \setlength{\abovedisplayshortskip}{0pt}%
  \setlength{\belowdisplayshortskip}{3pt}%
  \setlength{\jot}{1.5pt}%
}

\usepackage{hyperref}
\usepackage{url}

\title{ACCT: Asymmetric Cross-LLM\\Context Transfer}

\renewcommand{\thefootnote}{\fnsymbol{footnote}}

\author{
\textbf{Yuzhe Liu}\textsuperscript{1}\thanks{These authors contributed equally to this work.}
\quad
\textbf{Haolin Tian}\textsuperscript{1}\footnotemark[1]
\quad
\textbf{Tonghan Wang}\textsuperscript{1,\ensuremath{\dagger}}
\\
\textsuperscript{1}Tsinghua University, Beijing, China
\\
\textsuperscript{\ensuremath{\dagger}}Correspondence to:
\texttt{thw@mail.tsinghua.edu.cn}
}

\iclrfinalcopy 
\begin{document}

\maketitle
\lhead{}
\renewcommand{\thefootnote}{\arabic{footnote}}

\begin{abstract}
In real-world deployments, large language model (LLM) systems are increasingly evolving from single-model invocation toward heterogeneous deployments in which different models are dynamically invoked for inference. When the same context is reused across models, however, each newly selected target model typically has to process it again, repeatedly incurring substantial computational costs. Therefore, we propose ACCT (Asymmetric Cross-LLM Context Transfer), which targets KV cache transfer between models of different scales within the same model family. ACCT transforms the KV cache produced by a source model into sparse KV states that can be directly consumed by a target model, thereby enabling cross-model reuse of context computation and decoupling context computation from downstream inference. ACCT consists of three lightweight components: the Cartographer identifies query-relevant context tokens for each target layer and KV head; the Allocator determines how many context tokens each target layer
and KV head retains under a shared total KV budget; and the Compiler transforms selected source KV into sparse target KV. The target model can therefore perform inference using the transferred KV without reprocessing the original context. We evaluate ACCT in two complementary settings comprising five general-capability benchmarks and six highly context-dependent tasks. Experiments show that ACCT matches or exceeds the target model's original inference performance even when reducing the KV cache size by 75\%. At a context length of $\sim$19K tokens, ACCT's KV conversion achieves a $6.5$--$21.2\times$ speedup over target context prefill. 
These results demonstrate the feasibility of transferring context computation across models under a limited KV budget.
\end{abstract}

\section{Introduction}
\label{sec:introduction}

Large language model (LLM) systems are evolving from serving a single model to
heterogeneous deployments comprising models of different scales and capabilities~\citep{chen2024frugalgpt,liu2024droidspeak}.
Such systems may dynamically route requests according to task difficulty, latency requirements, and resource availability~\citep{ong2025routellm}.
Meanwhile, conversation
histories, retrieved documents, and task states often persist across requests,
far outliving any individual model invocation~\citep{packer2023memgpt,gim2024promptcache}. Thus, while the executing model may change,
the context often remains the same. Yet when the same context is handed to a new target model,
the target typically must process the entire context again to construct the internal states required
for inference. As contexts grow and are used more frequently across models, the
costs of repeated computation and state construction accumulate. Enabling models to reuse context computation performed by another model,
is therefore an
important challenge for efficient heterogeneous LLM systems.

Standard caching mechanisms reduce repeated computation within a model but do not directly make its KV cache reusable by other models.
KV caching reuses states for preceding tokens during generations, while prefix caching extends this reuse across requests served by the same model~\citep{kwon2023pagedattention,zheng2023sglang}. These states,
however, depend jointly on the model's parameters and architectures and are generally not interchangeable across models, even when the models belong to the same family~\citep{liu2024droidspeak,fu2025c2c}. 
Allowing a target model to build on context computation already performed
by a source model requires transforming source KV into KV that the target can read directly. The transfer process should
organize information around the target's reasoning needs, retaining content
relevant to the current query under a specified budget, and making it accessible by the target.

To this end, we propose Asymmetric Cross-LLM Context Transfer (ACCT) for
reusing context computation between models of different scales within the same
family. ACCT transfers KV 
already produced by the source to sparse KV that the
target can consume directly. To reuse source context computation under a limited
KV capacity, ACCT achieves information selection for the target, KV budget
allocation, and target KV generation through three lightweight components.
The \textbf{Cartographer} uses source context and query representations to rank content tokens for every target layer and KV head. The
\textbf{Allocator} assigns capacity across
layers and heads under a global budget; together with the content token rankings,
it determines which KV is retained. The \textbf{Compiler}
converts selected source KV into KV for the corresponding
target layers and KV heads. 
After mounting the transferred KV, the target
reads context information through its own attention mechanism, without
re-encoding the raw context, and performs only query encoding and subsequent
autoregressive generation. This interface integrates context computation already
performed by the source into the target's regular inference process, allowing
context encoding and downstream reasoning to be carried out by different models.

ACCT aims to preserve the target's reasoning capabilities while avoiding its
re-encoding of the raw context. We evaluate ACCT on five general-capability
benchmarks and six highly context-dependent tasks across six directed model
pairs. Using \emph{Target Full}, in which the target model processes the complete
context, as the reference, ACCT closely matches its performance on many
tasks and surpasses it on some. At a context length of approximately 19K tokens, ACCT's KV
conversion achieves a $6.5$--$21.2\times$ speedup over target context prefill. These results show that a target
can use context information through transferred KV to support subsequent
reasoning even without seeing the raw context itself. ACCT thus demonstrates
the feasibility of treating context computation as a resource that can be reused
across models, opening new possibilities for independently organizing context
processing and reasoning in heterogeneous model systems.


\section{Related Work}
\label{sec:related-work}

\textbf{Heterogeneous LLM Serving.}\ \
FrugalGPT~\citep{chen2024frugalgpt} constructs model cascades to reduce inference
costs, while RouteLLM~\citep{ong2025routellm} learns routing policies between
models with different capabilities and costs. These approaches focus on
selecting the executing model; ACCT complements them by enabling reuse of
previously computed context representations when execution switches to a
different-scale model.

\textbf{Context Reuse in LLM Inference.}\ \
Prompt Cache~\citep{gim2024promptcache} reuses attention states across requests,
while H$_2$O~\citep{zhang2023h2o}, SnapKV~\citep{li2024snapkv}, and
PyramidKV~\citep{cai2024pyramidkv} reduce retained KV states through token
selection or layer-wise budgeting. These methods reuse or compress states
within the model that produced them. Extending reuse across models,
DroidSpeak~\citep{liu2024droidspeak} shares caches between same-architecture
variants of a common base model and recomputes critical context layers in the
receiver. ACCT instead converts representations across model scales without
target-side context recomputation. Beyond computational reuse,
C2C~\citep{fu2025c2c} supports semantic communication across model sizes and
families by projecting and fusing source KV with the receiver's own context KV,
avoiding intermediate text generation. Its full method retains receiver-side
context encoding, whereas ACCT constructs the target's context memory from
source representations under a global state budget and runs the target only
on the query and subsequent generation. Accordingly, neither method in its original formulation provides a matched baseline for our setting, which requires cross-scale context-state transfer without target-side context encoding.

\section{Methods}
\label{sec:Methods}
ACCT converts context representations already computed by a source model into sparse target KV, allowing the target model to answer queries without re-encoding the full context.
\suppressfloats[t]
\begin{figure}[t]
\centering
\includegraphics[width=\linewidth]{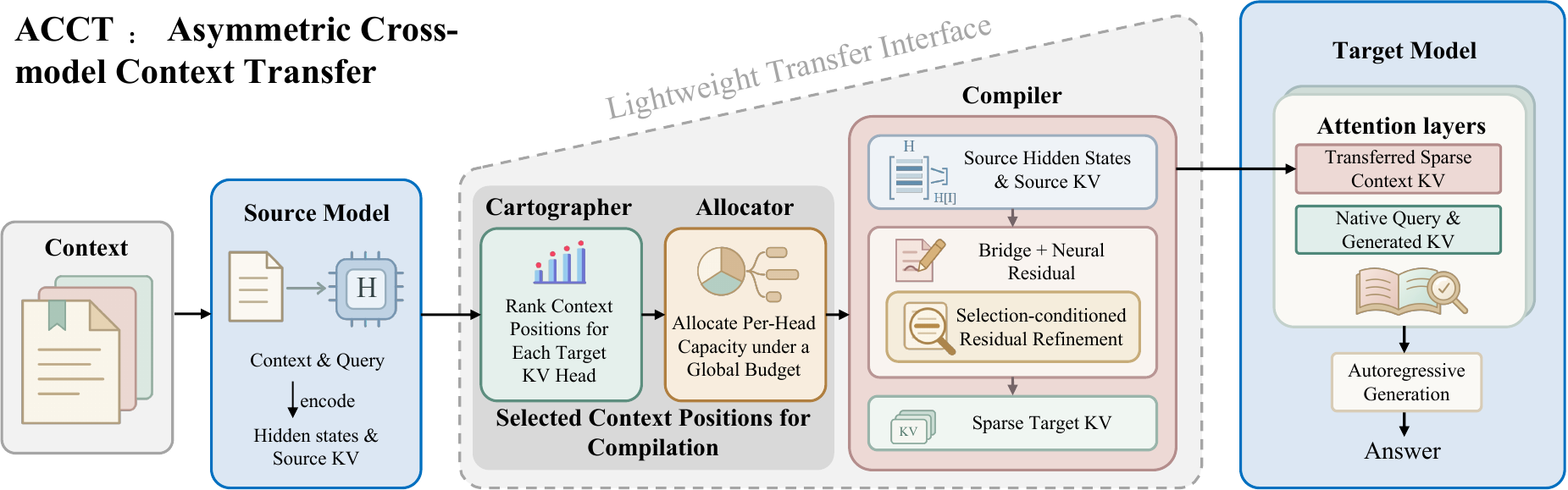}
\caption{ACCT combines Cartographer selection, Allocator budgeting, and Compiler conversion to transfer source KV into sparse target KV for inference without context re-encoding.}
\label{fig:acct-overview}
\end{figure}
\subsection{Transfer objective and overall workflow}
Consider frozen source and target models $F_S,F_T$ from the same model family, with a context $c$ containing $N$ tokens and a query $q$ containing $Q$ tokens. The source model processes the concatenated context and query to obtain final-layer context and query hidden states $H_c,H_q$, selected intermediate-layer context states $\{H_c^{(r)}\}_{r\in\mathcal R}$, and context KV. $\mathcal R$ indexes the selected source intermediate layers. The target has $L_T$ layers and $H_T$ KV heads per layer, each storing $d_T$-dimensional keys and values. We call KV head $h$ in layer $\ell$ a target unit $p=(\ell,h)$, giving $P=L_TH_T$ units.

We collectively call the three modules that convert source representations into sparse target KV the transfer interface: Cartographer scores context tokens for each unit using context and query information; Allocator assigns retention capacities under a global budget (the total number of tokens allowed across all target units) and selects by score; Compiler converts only selected source representations into target-readable keys and values. The budget counts retained context tokens across physical KV heads; in grouped-query attention, query heads sharing one KV head are not counted separately. Let $I_p$ be the set of original indices retained by unit $p$, $i$ an index in this set, and $\widehat K_{p,i},\widehat V_{p,i}$ the corresponding Compiler outputs. Denote the transferred context cache constructed from $c,q$ for the target by $\widehat{\mathcal C}_{S\to T}(c,q)$, where $S\to T$ indicates source-to-target transfer:
\begin{equation}
\widehat{\mathcal C}_{S\to T}(c,q)
=\left\{\left\{(i,\widehat K_{p,i},\widehat V_{p,i}):i\in I_p\right\}\right\}_{p=1}^{P}.
\end{equation}
The cache preserves original token indices for positional encoding. After loading these KV pairs, the target only processes the query and generates the answer.

\subsection{Constructing sparse target KV from source representations}

\subsubsection{Cartographer: scoring context tokens for each KV head}
Different target units serve different roles, so a token may differ in importance across units; its importance also depends on the query and its context position. Cartographer therefore combines source hidden states $H_c,H_q$, target layer and KV-head information, and token position to score context token $i$ in unit $p$ as $S_{p,i}$, establishing retention priorities within each unit. Specifically,
\begin{equation}
S_{p,i}=f_o(\gamma_p\odot x_i+\beta_p+W_m m_{p,i}+W_\phi\phi_i).
\end{equation}
Cartographer first extracts query-relevant content shared across units to avoid repeated query--context interaction. 
$x_i$ is a token vector extracted and projected from $H_c$. $\gamma_p,\beta_p$ are scaling and shifting vectors derived from the unit and query-relevant content. $m_{p,i}$ computes matching scores between token $i$ and that content conditioned on identification features of the target layer and KV head. The two-dimensional vector $\phi_i$ records token $i$'s normalized position in the context and its normalized log-distance from the context end. Learned matrices $W_m,W_\phi$ project matching and positional features to the token-vector dimension, $f_o$ maps the combined vector to a scalar, and $\odot$ denotes elementwise multiplication. Feature computations and the shared architecture are detailed in Appendix A.1.

\subsubsection{Allocator: assigning per-head capacities under a fixed budget}
Target units may need different amounts of context, which uniform allocation under a fixed budget cannot accommodate. Allocator therefore learns a capacity for each unit, a trainable parameter controlling its retention count, and determines the retained tokens based on Cartographer's ranking. 

To prevent context immediately preceding the query from being entirely discarded by ranking or capacity competition, each unit reserves the final $\tau$ context tokens and allocates the remaining capacity to earlier context, and the prefix contains $N-\tau$ tokens. Given retention ratio $\rho$, the budget allows an average of $k=\max(\tau,\lceil\rho N\rceil)$ context tokens per target unit, including the reserved tail. $\lceil\cdot\rceil$ rounds upward. Subtracting the reserved $P\tau$ from total capacity $Pk$ leaves a global budget of $P(k-\tau)$ for the prefix. Allocation must respect both prefix length and the total budget. Let $\alpha_p$ be unit $p$'s learned capacity and $b_p$ its real-valued prefix capacity before rounding. The function $\sigma(x)=1/(1+e^{-x})$ bounds the capacity fraction between zero and one, while a shared adjustment $\lambda$ enforces the total budget:
\begin{equation}
b_p=(N-\tau)\sigma(\alpha_p-\lambda),\qquad
\sum_{p=1}^{P}b_p=P(k-\tau).
\end{equation}
After converting $b_p$ to integer prefix capacities $k_p$ while preserving the total budget, each unit selects its $k_p$ highest-scoring prefix tokens and adds the final $\tau$ tokens to form $I_p$, retaining $k_p+\tau$ tokens in total. Answer prediction supervises the learning of $\alpha_p$ (Section 3.3). Allocation details are in Appendix C.

\subsubsection{Compiler: converting selected source representations into target KV}
The retained set $I_p$ determines which context tokens supply KV to target unit $p$, but their source KV cannot be used directly by the target model. Compiler therefore converts selected source KV into target KV, using final-layer states $H_c$ and three selected intermediate-layer states $\{H_c^{(r)}\}_{r\in\mathcal R}$ for context-dependent corrections. For each target unit, Bridge applies a linear transformation with bias to the selected tokens' source KV from a predetermined source layer and KV head, producing $K^B_{p,i},V^B_{p,i}$. 
A correction network predicts context-dependent corrections $\Delta K_{p,i},\Delta V_{p,i}$ based on the context. Adding them gives the result:
\begin{equation}
\widehat K_{p,i}=K^B_{p,i}+\Delta K_{p,i},
\qquad
\widehat V_{p,i}=V^B_{p,i}+\Delta V_{p,i}.
\end{equation}
The correction network reads the final-layer source states $H_c$ and the three selected intermediate-layer state sequences $\{H_c^{(r)}\}_{r\in\mathcal R}$, sharing the context computation and each selected token's representation across target units:
\begin{equation}
(\Delta K_{p,i},\Delta V_{p,i})
=\operatorname{Correction}_{\psi}\!\left(H_c,\{H_c^{(r)}\}_{r\in\mathcal R},\delta_p,i;\,p\right).
\end{equation}
Here, $\delta_p=|I_p|/N$ is the retained fraction, $\psi$ is trainable parameters.
Source-layer/head correspondence, the correction architecture, positional-encoding handling, and fitting details are in Appendix A.2.

\subsection{The Training Scheme}
We train the three components in stages. We first learn scoring and allocation with target-native KV, which is keys and values produced by the target itself when encoding context. We then fix the retained token sets and train Compiler to convert the selected source representations into the corresponding target KV. Only the transfer interface is updated in these stages, leaving the source and the target model intact.


\textbf{Scoring: learn to rank by the effect of token removal.}
To retain useful tokens, our supervision reflects the effect of discarding them. We remove a context token's key and value while keeping the target model's query vectors and the remaining keys and values fixed, then renormalize the remaining attention weights. A larger change in the attention output indicates a more important token. We collect this change at the query end and on subsequent tokens generated by the target for supervision, aggregating it per KV head as deletion utility $u_{p,i}$.

To train Cartographer before learning the Allocator, we temporarily assign equal retention capacity to every target unit. Within each unit, we rank prefix tokens by deletion utility; the boundary between the last retained token and the first discarded token defines the retention cutoff. Tokens on opposite sides of the retention cutoff often have similar deletion utilities, making them the hardest cases for the Cartographer to distinguish. We therefore sample training pairs near the retention cutoff. Let $\mathcal P$ denote these pairs, each containing a higher-utility token $i$ and a lower-utility token $i'$, with $u_{p,i}>u_{p,i'}$. The ranking loss encourages $S_{p,i}>S_{p,i'}$. For clarity, we present the loss without score normalization; the implemented normalization and pair-sampling scheme are detailed in Appendix B.1:
\begin{equation}
\mathcal L_{\mathrm{rank}}=-\frac{1}{|\mathcal P|}
\sum_{(p,i,i')\in\mathcal P}
\log\sigma\!\left(S_{p,i}-S_{p,i'}\right).
\end{equation}

\textbf{Allocation: learn to use limited capacity to support answer prediction.}
We then freeze Cartographer and adjust capacity preferences using the target's prediction of training answers. With the target model frozen, we optimize the capacity preferences to increase the probability of the next correct answer token given the preceding correct answer tokens and target-native KV of the context. Let $\alpha$ denote the capacity preferences of all target units, and let $I(\alpha)=(I_p(\alpha))_{p=1}^{P}$ denote their retained token-index sets. The cache $\mathcal C_T(c)|_{I(\alpha)}$ contains each unit's target-native KV at its retained indices. Let $y$ be the correct answer containing $|y|$ tokens, with $y_j$ denoting token $j$ and $y_{<j}$ the preceding correct answer tokens; $p_T$ denotes the target model's prediction probability. The allocation loss is
\begin{equation}
\mathcal L_{\mathrm{alloc}}
=-\frac{1}{|y|}\sum_{j=0}^{|y|-1}
\log p_T\!\left(
y_j\mid q,y_{<j};\mathcal C_T(c)|_{I(\alpha)}
\right).
\end{equation}
Since rounding and discrete token selection are non-differentiable, to learn $\alpha$, we introduce a soft-gate threshold that varies continuously with capacity $b_p$, and $b_p$ is differentiable with respect to the capacity preferences $\alpha$. Therefore, gradients from the answer-prediction loss pass through the soft gates to $b_p$, and then, by the chain rule, to $\alpha$, updating the capacity preferences.
The correct answer $y$ is not provided to Cartographer or Allocator. Implementation details are given in Appendix C.

\textbf{Conversion: learn to transform KV.}
We fix the retained token sets, fit Bridge using paired source and target-native KV for the same tokens, then freeze Bridge and train the correction network so that the converted KV approximates the corresponding target-native KV. Converted KV should match native KV and support similar attention and predictions, motivating losses at three levels. At the representation level, $\mathcal L_{K,\mathrm{norm}}$ compares the converted and target-native key vectors after root-mean-square normalization, while $\mathcal L_{K,\mathrm{scale}}$ compares the logarithms of their root-mean-square magnitudes. $\mathcal L_V$ compares the directions and lengths of the converted and target-native value vectors.
At the attention level, $\mathcal L_{K,\mathrm{attn}}$ compares attention distributions with target query vectors fixed and the selected target-native context keys replaced by their converted counterparts. At the prediction level, $\mathcal L_{\mathrm{beh}}$ compares the target model's next-token distributions when the same query tokens are processed with the selected target-native KV and the converted KV. The query states are recomputed under each cache. The total loss is
\begin{equation}
\mathcal L_{\mathrm{comp}}=
\mathcal L_{K,\mathrm{norm}}+10^{-4}\mathcal L_{K,\mathrm{scale}}
+\mathcal L_{K,\mathrm{attn}}+\mathcal L_V+\mathcal L_{\mathrm{beh}}.
\end{equation}
Normalization, distribution comparisons, and numerical stabilizers are detailed in Appendix B.

\subsection{Inference and computational cost}
Inference replaces target context encoding with transferred context KV. The source first processes context and query; the interface scores, allocates, and converts, then writes the resulting KV into the target model's cache. The target reads this cache, processes only the query, and generates the answer. The target model computes and appends the KV for the query and newly generated tokens as usual. Sparse selection reduces context KV storage: each token retained by a KV head contributes two $d_T$-dimensional vectors, so capacity $Pk$ requires $2Pkd_T$ scalars, while full context requires $2PNd_T$ scalars, leading to a ratio of $k/N$.
The full inference algorithm and complexity analysis are in Appendix F.2.

\section{Experiments}
\label{sec:experiments}
We evaluate ACCT in both transfer directions across three model families,
examining whether transferred states support target inference and how state
capacity affects different tasks. Although these target models also serve as source models in the reverse-direction pairs, we do not directly compare their results with those of the reverse-direction pairs here, because the two settings use different decoding models, and differences in task performance cannot be attributed solely to state transfer.

\subsection{Experimental Setup}
\label{sec:experiment-setup}

\textbf{Models and datasets.}\ \
We evaluate bidirectional transfer between Qwen3-4B and Qwen3-8B,
Llama-3.2-3B and Llama-3.1-8B, and Qwen2.5-7B and Qwen2.5-32B,
yielding six directed settings.
Each pair shares a token-to-ID mapping and uses the target tokenizer and prompts.

Evaluation covers five general-capability benchmarks (HellaSwag, Winogrande,
MMLU, PIQA, and WikiText2) and six context-dependent tasks (SearchQA, TriviaQA,
BoolQ, RACE, WikiHop, and PubMedQA). We use 500 examples each from WikiText2
and PubMedQA and 1,000 from each remaining dataset, yielding the same fixed
set of 10,000 evaluation examples for every direction.

\textbf{Comparison setup.}\ \
We evaluate the reasoning performance supported by transferred states when
the target does not re-encode the context, together with the computational
cost of constructing these states. Target Full provides a task-performance
reference by processing the complete context natively, whereas ACCT supports
target inference through cross-model state transfer. We assess this process
in terms of both task performance and state-construction latency. DroidSpeak requires models with the same architecture, whereas C2C relies on the target model’s own context prefill; neither directly satisfies our cross-scale setting without target-side context re-encoding, so we exclude them from the main-table comparison.

\textbf{Training setup.}\ \
We train a separate interface for each transfer direction while keeping both
source and target models frozen. Main experiments use $4\times$ compression,
retaining approximately one quarter of target context KV slots under the budget
rule in Section~3, including the most recent 32 positions. The Cartographer
and Allocator use 5,000 SearchQA training examples; the Compiler uses a fixed
subset of 500 for four epochs. Training schedules, model sources, and interface
capacity rules are provided in Supplementary Material D--E.

\textbf{Evaluation metrics.}\ \ 
We report token-level F1 for SearchQA and TriviaQA, PPL computed from mean
token negative log-likelihood for WikiText2, and accuracy for the remaining
tasks. General multiple-choice tasks use candidate-answer likelihoods;
dataset sources and evaluation details appear in Appendix G. F1 and accuracy
are reported as percentages (higher is better); PPL is reported on its original
scale (lower is better). For efficiency, we compare KV state-conversion latency
with native target context prefill latency; the measurement protocol is given
in Appendix~\ref{app:latency-protocol}.

\subsection{Inference Efficiency}
\label{sec:inference-efficiency}
\textbf{Longer contexts amplify reuse gains.}\ \
Figure~\ref{fig:context-reuse-latency} shows faster state conversion in all six
directions at every measured length. Speedups increase from $1.5$--$8.6\times$
on the shortest contexts (approximately 1K tokens) to $6.5$--$21.2\times$ on the
longest (approximately 19K tokens), where conversion requires only
$4.7$--$15.3\%$ of target prefill time.
Target prefill processes every context token through the full target network,
whereas ACCT constructs target states from existing source representations
using shared lightweight computation. Conversion latency grows more slowly
with context length, making the savings from avoiding target re-encoding
increasingly pronounced.

\begingroup
\setlength{\intextsep}{3pt}
\begin{figure}[htbp]
\centering
\setlength{\abovecaptionskip}{2pt}
\includegraphics[width=0.80\linewidth]{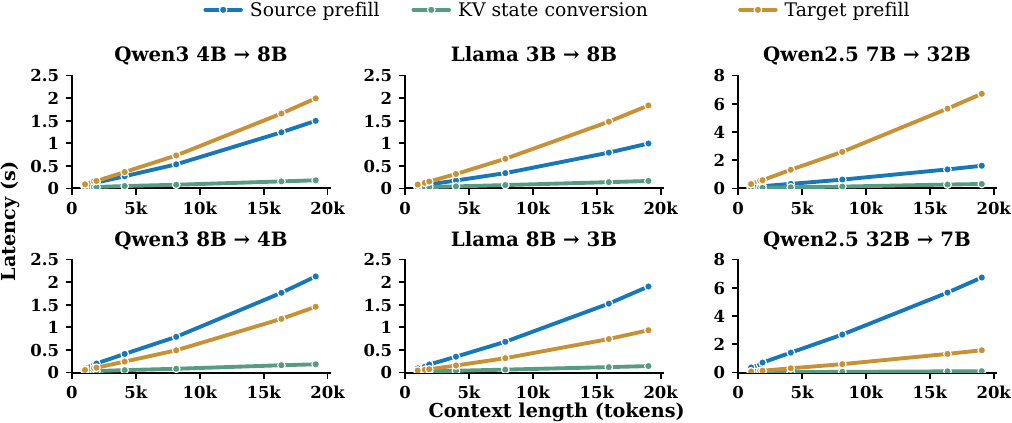}
\caption{Context-state construction latency across six transfer directions. Target prefill encodes the complete context with the target model, whereas ACCT converts already computed source states. The comparison shows the efficiency gain from avoiding target-side context re-encoding, especially for longer contexts.}
\label{fig:context-reuse-latency}
\end{figure}
\endgroup

\FloatBarrier
\subsection{Task Performance}
\label{sec:main-results}
Table~\ref{tab:main-results} summarizes task performance across three model
families in both directions at $4\times$ compression. Alongside the substantial
reduction in target context computation, ACCT incurs only modest performance
losses on several tasks and exceeds Target Full on others.

\begin{table}[t]
\centering
\footnotesize
\setlength{\tabcolsep}{1.2pt}
\renewcommand{\arraystretch}{1.10}
\caption{Task performance of ACCT at $4\times$ compression versus Target Full, which uses the target model's full context KV. Even with 4$\times$ compression (75\% reduction in the global cache usage) and up to 95.3\% reduction in prefilling time, our method can outperform full target KV on several tasks. F1 and accuracy are percentages (higher is better); PPL is lower-is-better. Bold marks improvements over Target Full. Relative is ACCT/Target Full for F1 and accuracy, and Target Full/ACCT for PPL.}
\label{tab:main-results}
\begin{tabular*}{\textwidth}{@{\extracolsep{\fill}}ll*{11}{r}@{}}
\toprule
& & \multicolumn{5}{c}{General capability} & \multicolumn{6}{c}{Context-dependent tasks} \\
\cmidrule(lr){3-7}\cmidrule(lr){8-13}
Model pair & Method & \shortstack{Hella\\Swag} & \shortstack{Wino\\grande} & MMLU & PIQA & \shortstack{Wiki\\Text2} & \shortstack{Search\\QA} & \shortstack{Trivia\\QA} & BoolQ & RACE & WikiHop & \shortstack{PubMed\\QA} \\
 & & Acc. & Acc. & Acc. & Acc. & PPL $\downarrow$ & F1 & F1 & Acc. & Acc. & Acc. & Acc. \\
\midrule
\multirow{3}{*}{\shortstack[l]{Qwen3\\$4\mathrm{B}\!\to\!8\mathrm{B}$}} & Target Full & 39.70 & 58.50 & 59.70 & 73.20 & 32.91 & 46.65 & 50.48 & 86.30 & 87.60 & 49.60 & 48.80 \\
 & ACCT & 33.20 & 51.50 & \textbf{65.70} & 73.10 & \textbf{25.94} & \textbf{67.07} & \textbf{66.31} & 61.90 & 63.50 & 36.50 & 40.20 \\
 & Relative & $0.84\times$ & $0.88\times$ & $1.10\times$ & $1.00\times$ & $1.27\times$ & $1.44\times$ & $1.31\times$ & $0.72\times$ & $0.72\times$ & $0.74\times$ & $0.82\times$ \\
\midrule
\multirow{3}{*}{\shortstack[l]{Qwen3\\$8\mathrm{B}\!\to\!4\mathrm{B}$}} & Target Full & 37.10 & 51.00 & 63.60 & 69.20 & 81.29 & 38.56 & 37.68 & 84.00 & 87.10 & 45.80 & 34.00 \\
 & ACCT & 36.90 & \textbf{51.30} & 58.20 & 69.00 & 98.92 & \textbf{50.33} & \textbf{54.51} & 60.20 & 66.30 & 32.50 & \textbf{49.60} \\
 & Relative & $0.99\times$ & $1.01\times$ & $0.92\times$ & $1.00\times$ & $0.82\times$ & $1.31\times$ & $1.45\times$ & $0.72\times$ & $0.76\times$ & $0.71\times$ & $1.46\times$ \\
\midrule
\multirow{3}{*}{\shortstack[l]{Llama\\$3\mathrm{B}\!\to\!8\mathrm{B}$}} & Target Full & 54.60 & 49.10 & 58.60 & 77.40 & 7.21 & 63.44 & 78.71 & 82.00 & 68.30 & 37.90 & 70.00 \\
 & ACCT & 27.00 & \textbf{50.10} & 41.90 & 69.30 & 64.91 & \textbf{66.88} & 69.71 & 63.50 & 46.20 & 18.60 & 53.60 \\
 & Relative & $0.49\times$ & $1.02\times$ & $0.72\times$ & $0.90\times$ & $0.11\times$ & $1.05\times$ & $0.89\times$ & $0.77\times$ & $0.68\times$ & $0.49\times$ & $0.77\times$ \\
\midrule
\multirow{3}{*}{\shortstack[l]{Llama\\$8\mathrm{B}\!\to\!3\mathrm{B}$}} & Target Full & 50.20 & 49.30 & 45.20 & 73.80 & 9.04 & 37.96 & 69.50 & 69.00 & 57.60 & 29.60 & 66.00 \\
 & ACCT & 27.00 & \textbf{50.50} & 33.00 & 68.70 & 65.36 & \textbf{54.00} & 56.45 & 42.60 & 32.20 & 16.80 & 33.20 \\
 & Relative & $0.54\times$ & $1.02\times$ & $0.73\times$ & $0.93\times$ & $0.14\times$ & $1.42\times$ & $0.81\times$ & $0.62\times$ & $0.56\times$ & $0.57\times$ & $0.50\times$ \\
\midrule
\multirow{3}{*}{\shortstack[l]{Qwen2.5\\$7\mathrm{B}\!\to\!32\mathrm{B}$}} & Target Full & 61.10 & 57.80 & 72.40 & 80.30 & 5.83 & 44.05 & 34.10 & 85.30 & 93.90 & 50.00 & 57.00 \\
 & ACCT & 29.60 & 50.10 & 64.60 & 70.20 & 26.24 & \textbf{46.47} & 33.58 & 56.60 & 57.50 & 25.40 & 47.00 \\
 & Relative & $0.48\times$ & $0.87\times$ & $0.89\times$ & $0.87\times$ & $0.22\times$ & $1.05\times$ & $0.98\times$ & $0.66\times$ & $0.61\times$ & $0.51\times$ & $0.82\times$ \\
\midrule
\multirow{3}{*}{\shortstack[l]{Qwen2.5\\$32\mathrm{B}\!\to\!7\mathrm{B}$}} & Target Full & 57.00 & 51.80 & 54.30 & 76.80 & 8.40 & 46.63 & 41.15 & 84.60 & 85.30 & 33.00 & 64.60 \\
 & ACCT & 38.50 & 49.30 & 50.90 & 69.10 & 26.35 & 22.35 & 25.48 & 63.00 & 47.30 & 18.50 & 54.80 \\
 & Relative & $0.68\times$ & $0.95\times$ & $0.94\times$ & $0.90\times$ & $0.32\times$ & $0.48\times$ & $0.62\times$ & $0.74\times$ & $0.55\times$ & $0.56\times$ & $0.85\times$ \\
\bottomrule
\end{tabular*}
\end{table}

\textbf{Compact states effectively support target inference.}\ \
ACCT partially preserves capabilities across diverse tasks through compressed source
representations, with Qwen3 approaching or exceeding Target Full on multiple
benchmarks. These results show that the target can use context information
through transferred states without re-encoding the original context, connecting
existing context computation to downstream inference.

\textbf{Transfer capabilities extend across tasks.}\ \
The interface uses only SearchQA for training, yet Qwen3 transfer also yields
improvements over Target Full on TriviaQA, MMLU, WikiText2, and PubMedQA, spanning open-domain QA,
knowledge reasoning, language modeling, and domain-specific QA. All six
directions also retain much of the target's PIQA performance, demonstrating
the potential for task generalization from a single training dataset.

\textbf{The method's interface design is reusable across model families.}\ \
We apply the same three-component architecture and capacity adaptation rules to within-family model pairs in Qwen3, Llama, and Qwen2.5, training a separate interface for each transfer direction. These experiments demonstrate the applicability of a shared interface design across model families, while context transfer itself is evaluated between models within the same family.

\textbf{Transfer effectiveness depends on the task.}\ \
Gains are more pronounced in open-domain QA, while gaps from Target Full
remain in reading comprehension, multi-hop reasoning, and some language-modeling
settings. This suggests that task demands for context coverage and representation
precision influence transfer.

\subsection{Fine-Grained Analysis}
\label{sec:compression-analysis}
Table~\ref{tab:compression-results} compares compression budgets for Qwen3
$4\mathrm{B}\!\leftrightarrow\!8\mathrm{B}$. Each setting is trained separately,
with Compiler width, depth, and training schedule fixed within each direction.

\textbf{The capacity budget offers flexible control over context use.}\ \
Moving from $4\times$ to $2\times$ compression improves most tasks in both
directions, including all six context tasks in reverse transfer. The budget
thus provides a practical way to adjust how much context information ACCT
retains and choose state capacity for different inference needs.

\textbf{Information selection makes compact states more effective.}\ \
The $1\times$ setting jointly ablates the Cartographer and Allocator while
retaining the Compiler. On open-domain QA in both directions, selective
transfer at $2\times$ and $4\times$ outperforms full retention at $1\times$,
showing that, on these open-domain QA tasks, selective transfer achieves
better performance than full retention with fewer KV states.

\begingroup
\setlength{\floatsep}{6pt}
\setlength{\textfloatsep}{8pt}
\setlength{\intextsep}{8pt}
\begin{table}[!t]
\centering\fontsize{8.5}{9}\selectfont
\setlength{\abovecaptionskip}{4pt}
\setlength{\belowcaptionskip}{0pt}
\setlength{\tabcolsep}{1.2pt}
\renewcommand{\arraystretch}{1.00}
\caption{Task performance of bidirectional Qwen3 4B$\leftrightarrow$8B transfer at $1\times$ and $2\times$ compression, compared with the $4\times$ setting used in the previous experiments. 
ACCT supports target inference at different compression ratios and outperforms full retention on multiple tasks with fewer KV states.}
\label{tab:compression-results}
\begin{tabular*}{\textwidth}{@{\extracolsep{\fill}}ll*{11}{r}@{}}
\toprule
Model pair & Compression & \shortstack{Hella\\Swag} & \shortstack{Wino\\grande} & MMLU & PIQA & \shortstack{Wiki\\Text2} & \shortstack{Search\\QA} & \shortstack{Trivia\\QA} & BoolQ & RACE & WikiHop & \shortstack{PubMed\\QA} \\
 & & Acc. & Acc. & Acc. & Acc. & PPL $\downarrow$ & F1 & F1 & Acc. & Acc. & Acc. & Acc. \\
\midrule
\multirow{3}{*}{\shortstack[l]{Qwen3\\$4\mathrm{B}\!\to\!8\mathrm{B}$}} & $1\times$ & 38.60 & 51.40 & 55.80 & 70.90 & 23.26 & 38.54 & 47.03 & 77.80 & 69.40 & 35.10 & 45.60 \\
 & $2\times$ & 36.80 & 53.10 & 62.70 & 73.20 & 22.16 & 67.59 & 74.92 & 70.30 & 71.10 & 39.10 & 27.60 \\
 & $4\times$ & 33.20 & 51.50 & 65.70 & 73.10 & 25.94 & 67.07 & 66.31 & 61.90 & 63.50 & 36.50 & 40.20 \\
\midrule
\multirow{3}{*}{\shortstack[l]{Qwen3\\$8\mathrm{B}\!\to\!4\mathrm{B}$}} & $1\times$ & 37.40 & 50.50 & 44.70 & 69.30 & 60.74 & 12.54 & 14.11 & 68.00 & 56.00 & 4.80 & 48.40 \\
 & $2\times$ & 32.80 & 52.50 & 53.60 & 70.10 & 142.09 & 61.54 & 57.65 & 62.60 & 66.80 & 36.10 & 55.20 \\
 & $4\times$ & 36.90 & 51.30 & 58.20 & 69.00 & 98.92 & 50.33 & 54.51 & 60.20 & 66.30 & 32.50 & 49.60 \\
\bottomrule
\end{tabular*}
\end{table}

\textbf{Component ablations and state controls.}\ \
At $4\times$ compression with the 32-token tail, \emph{Random} replaces
Cartographer rankings with random orderings and retrains the Allocator and Compiler; \emph{Uniform}
retains rankings, equalizes per-layer/head capacity, and retrains the Compiler.
At unchanged positions, \emph{Bridge-only} removes the neural residual and
\emph{Native KV} uses target-native states, both without retraining.
Table~\ref{tab:component-ablation} reports both directions; matched training
settings and the 80,000-example evaluation protocol are detailed in
Appendix~\ref{app:compiler-ablation}.

\begin{table}[htbp]
\centering\fontsize{8.5}{9}\selectfont
\setlength{\abovecaptionskip}{4pt}
\setlength{\belowcaptionskip}{0pt}
\setlength{\tabcolsep}{1.2pt}
\renewcommand{\arraystretch}{1.00}
\caption{Component ablations for Qwen3 transfer at $4\times$ compression. Random replaces learned selection with random selection, Uniform replaces learned allocation with equal allocation, Bridge-only removes the correction procedure, and Native KV uses target-native states at the same positions. Cartographer, allocator, compiler all contribute significantly to the performance of ACCT.}
\label{tab:component-ablation}
\begin{tabular*}{\textwidth}{@{\extracolsep{\fill}}ll*{11}{r}@{}}
\toprule
Model pair & Method & \shortstack{Hella\\Swag} & \shortstack{Wino\\grande} & MMLU & PIQA & \shortstack{Wiki\\Text2} & \shortstack{Search\\QA} & \shortstack{Trivia\\QA} & BoolQ & RACE & WikiHop & \shortstack{PubMed\\QA} \\
 & & Acc. & Acc. & Acc. & Acc. & PPL $\downarrow$ & F1 & F1 & Acc. & Acc. & Acc. & Acc. \\
\midrule
\multirow{5}{*}{\shortstack[l]{Qwen3\\$4\mathrm{B}\!\to\!8\mathrm{B}$}} & ACCT & 33.20 & 51.50 & 65.70 & 73.10 & 25.94 & 67.07 & 66.31 & 61.90 & 63.50 & 36.50 & 40.20 \\
 & Random & 31.70 & 50.80 & 27.00 & 61.20 & 68.42 & 11.09 & 20.12 & 38.90 & 27.30 & 15.30 & 51.60 \\
 & Uniform & 35.20 & 50.60 & 62.80 & 72.40 & 25.24 & 41.07 & 42.72 & 62.90 & 64.20 & 34.10 & 22.80 \\
 & Bridge-only & 35.10 & 51.10 & 41.70 & 70.20 & 35.24 & 44.15 & 50.13 & 63.10 & 61.70 & 33.80 & 15.40 \\
 & Native KV & 38.00 & 57.40 & 54.30 & 73.20 & 37.19 & 59.94 & 59.58 & 67.20 & 75.30 & 44.50 & 36.20 \\
\midrule
\multirow{5}{*}{\shortstack[l]{Qwen3\\$8\mathrm{B}\!\to\!4\mathrm{B}$}} & ACCT & 36.90 & 51.30 & 58.20 & 69.00 & 98.92 & 50.33 & 54.51 & 60.20 & 66.30 & 32.50 & 49.60 \\
 & Random & 29.20 & 50.00 & 23.50 & 57.30 & 172.93 & 3.67 & 5.20 & 38.90 & 24.90 & 3.70 & 19.40 \\
 & Uniform & 31.90 & 51.50 & 51.10 & 69.00 & 145.68 & 57.82 & 59.44 & 61.20 & 64.10 & 35.50 & 58.00 \\
 & Bridge-only & 31.30 & 48.50 & 41.20 & 64.80 & 155.10 & 36.28 & 34.98 & 49.60 & 58.80 & 31.90 & 17.40 \\
 & Native KV & 32.50 & 50.90 & 62.20 & 69.20 & 117.63 & 56.26 & 49.92 & 51.50 & 74.80 & 43.00 & 16.80 \\
\bottomrule
\end{tabular*}
\end{table}

\textbf{Learned localization is central to effective transfer.}\ \
Random selection substantially degrades diverse tasks in both directions
even after downstream components are retrained. Learned localization maintains
advantages in QA, knowledge reasoning, and language modeling, showing that
the Cartographer retains more useful content under a fixed budget and provides
an effective information basis for allocation and compilation.

\textbf{Learned allocation benefits multiple benchmarks.}\ \
At the same total budget, learned allocation improves performance
on multiple benchmarks, including forward open-domain QA and
MMLU in both directions, but shows no advantage over uniform
allocation on some benchmarks.

\textbf{Residual compilation strengthens cross-model state alignment.}\ \
The full Compiler improves open-domain QA and MMLU and lowers
language-modeling perplexity relative to Bridge-only in both directions;
reverse-transfer gains cover every evaluated task. The residual network
effectively complements affine mapping, making selected source representations
more suitable for target consumption.

\textbf{Compiled states approach or surpass native KV on multiple tasks.}\ \
With positions and capacity fixed, ACCT nearly matches Native KV on PIQA
and outperforms it on TriviaQA, PubMedQA, and WikiText2 in both directions.
This control evaluates the quality of Compiler-generated states, showing that
source representations can be compiled into target states with inference
performance approaching or exceeding native KV on these tasks.

\FloatBarrier

\section{Conclusion}

ACCT decouples context computation from inference by compiling source
representations into sparse target-native KV. Bidirectional experiments across
three model families show performance approaching or exceeding Target Full on
multiple benchmarks. KV state conversion substantially outpaces
target context prefill, especially for long contexts.
These findings support context computation as a reusable resource across models,
enabling flexible and efficient heterogeneous LLM systems.

\endgroup

\section*{AI Use Statement}
Generative AI tools were used to assist with literature search, manuscript
drafting and language refinement, code implementation and debugging,
experimental execution and reproduction, analysis of experimental results,
and manuscript formatting. The authors reviewed and revised the AI-assisted
text, code, and analyses, and take full responsibility for the research design,
experimental procedures, reported results, references, and final content of
this work, including all claims and artifacts produced with the aid of
generative AI.

\section*{Reproducibility Statement}
Section~\ref{sec:Methods} and Appendices~\ref{app:component-parameterization}--\ref{app:budget-derivations}
describe the architecture, supervision, and budget optimization.
Appendices~\ref{app:training-configuration}--\ref{app:execution-interface}
document training schedules, random seeds, model-pair capacity rules, and
execution interfaces. Section~\ref{sec:experiments} and
Appendix~\ref{app:data-evaluation} specify the datasets, input and decoding
protocols, and evaluation metrics; Appendix~\ref{app:compiler-ablation}
details the ablation controls. Appendix~\ref{app:latency-protocol} reports
hardware and execution settings, timing boundaries, repetition counts, and
per-length latency measurements, including state construction with source
encoding. Additional model-pair results are provided in
Appendix~\ref{app:additional-pairs}.

\bibliographystyle{iclr2027_conference}
\bibliography{iclr2027_conference}

\clearpage
\appendix
\renewcommand{\theequation}{\thesection.\arabic{equation}}

\FloatBarrier
\section{Component Parameterization and Bridge Fitting}
\label{app:component-parameterization}

\subsection{Cartographer: query-conditioned ranking}

Let $H_c\in\mathbb R^{N\times D_S}$ and $H_q\in\mathbb R^{Q\times D_S}$ denote final-layer source context and query states. A shared map $f_\theta$ applies LayerNorm~\citep{ba2016layernorm}, a linear projection, GELU~\citep{hendrycks2016gelu}, and LayerNorm, producing width-$w_M$ features $X=f_\theta(H_c)$ and $Y=f_\theta(H_q)$. Four learnable query slots $A\in\mathbb R^{4\times w_M}$ aggregate the query and then read context using scaled dot-product attention~\citep{vaswani2017attention} with $d_M$-dimensional projections:

\begin{equation}
U=\operatorname{softmax}(AY^\top/\sqrt{w_M})Y,\qquad
Z=\operatorname{LN}\!\left(UW_Q+\operatorname{softmax}\!\left(\frac{(UW_Q)(XW_K)^\top}{\sqrt{d_M}}\right)XW_V\right).
\label{app:eq:A1}
\end{equation}

The learnable matrices $W_Q,W_K,W_V\in\mathbb R^{w_M\times d_M}$ project the features. The two softmax operations normalize over query and context positions, respectively, excluding padding. All target units share $Z$. For $p=(\ell,h)$, concatenate $c_M$-dimensional layer and head embeddings and project them to $e_p\in\mathbb R^{d_M}$. Together with $\bar Z$, this gives FiLM parameters $(\widetilde\gamma_p,\beta_p)=\operatorname{MLP}([e_p;\bar Z])$, where $\gamma_p=1+\tanh\widetilde\gamma_p$.

\begin{equation}
\begin{aligned}
x_i&=W_xX_i+b_x,\qquad E_{p,a}=Z_a\odot[1+\tanh(W_ge_p+b_g)],\\
m_{p,i,a}&=\langle x_i,E_{p,a}\rangle/\sqrt{d_M},\qquad
\phi_i=\left[i/N,\ \log(1+N-i)/\log(1+N)\right],\\
S_{p,i}&=w_o^\top\operatorname{Dropout}\!\left(\operatorname{GELU}
(\gamma_p\odot x_i+\beta_p+W_m\mathbf m_{p,i}+W_\phi\phi_i)\right).
\end{aligned}
\label{app:eq:A2}
\end{equation}

Here $a\in\{1,\ldots,4\}$ indexes query slots, $\mathbf m_{p,i}$ collects their four matching scores, and $i\in\{0,\ldots,N-1\}$. The learnable $W_x,b_x$ project tokens, $W_g,b_g$ modulate slots, $W_m,W_\phi$ project matching and position features, and $w_o$ reads out a scalar. Inference disables dropout and stably sorts non-tail positions by decreasing score. Scores determine content order; the Allocator determines capacity separately.

\subsection{Compiler: affine Bridge and shared residual}

\subsubsection{Affine Bridge inside the Compiler}

The source has $L_S$ layers, $H_S$ physical KV heads, and head dimension $d_S$. Relative-depth alignment and a modulo head map give $\pi_L(\ell)=\lfloor\ell(L_S-1)/(L_T-1)+1/2\rfloor$ and $\pi_H(h)=h\bmod H_S$; set $\pi_L(0)=0$ if $L_T=1$. These are fixed structural correspondences, not assumed semantic equivalences. Each target unit has separate affine key and value maps:

\begin{equation}
K^B_{p,i}=K^S_{\pi_L(\ell),\pi_H(h),i}A^K_p+a^K_p,\qquad
V^B_{p,i}=V^S_{\pi_L(\ell),\pi_H(h),i}A^V_p+a^V_p.
\label{app:eq:A3}
\end{equation}

The matrices $A_p^{K/V}\in\mathbb R^{d_S\times d_T}$ have $d_T$-dimensional biases. Keys are taken after native normalization and before RoPE. Fitting uses only valid positions in the frozen selections. For one head and vector type, let $X,Y$ stack $n_p$ source and target observations, with centered matrices $X_c,Y_c$. Then

\begin{equation}
\begin{aligned}
\min_{A,a}\;&\|XA+\mathbf1a^\top-Y\|_F^2+\lambda_B(n_p/d_S)\|A\|_F^2,\\
A&=[X_c^\top X_c+\lambda_B(n_p/d_S)I]^{-1}X_c^\top Y_c,
\qquad a=\bar y-\bar xA.
\end{aligned}
\label{app:eq:A4}
\end{equation}

We accumulate sufficient statistics and solve the linear system in double precision, without penalizing the bias. Set $\lambda_B=10^{-2}d_T/128$. The Bridge is fitted on the Compiler's 500 training examples and frozen before optimizing the neural residual on the same subset.

\subsubsection{Shared residual and native KV mounting}

The residual network concatenates the final source layer and three intermediate layers at $\ell_j=\operatorname{clip}(\operatorname{round}(jL_S/4),1,L_S)$ for $j=1,2,3$, indexed in the returned hidden-state sequence. A linear projection gives $E\in\mathbb R^{N\times w_W}$. Following the latent read/write attention pattern of Perceiver IO~\citep{jaegle2022perceiver}, starting from $m_W$ learnable latent slots, each of $D_Z$ blocks applies slot self-attention, cross-attention to all of $E$, and a SwiGLU~\citep{shazeer2020swiglu} feedforward transformation, with RMSNorm~\citep{zhang2019rmsnorm} and residual connections at every sublayer.

\begin{equation}
\begin{aligned}
Z'&=Z+\operatorname{SA}(\operatorname{RMSNorm}(Z)),\\
Z''&=Z'+\operatorname{CA}(\operatorname{RMSNorm}(Z'),E),\\
Z^+&=Z''+\operatorname{SwiGLU}(\operatorname{RMSNorm}(Z'')),\qquad
\operatorname{SwiGLU}(x)=[\operatorname{SiLU}(xW_1)\odot xW_2]W_3.
\end{aligned}
\label{app:eq:A5}
\end{equation}

The feedforward matrices are $W_1,W_2\in\mathbb R^{w_W\times f_W}$ and $W_3\in\mathbb R^{f_W\times w_W}$; SA and CA have learnable attention projections. For each position in $\mathcal U=\bigcup_p\mathcal I_p$, initialize with $E_i$ and apply $D_U$ blocks containing cross-attention to latent slots and feedforward transformations, followed by RMSNorm to obtain $z_i$. Selected positions do not self-attend. All target units selecting the same position share its $z_i$.

\begin{equation}
\begin{aligned}
\xi_{p,i}&=(\delta_p,\delta_p^2,i/\max(1,N-1),1),\quad
\delta_p=|\mathcal I_p|/N,\quad \widetilde z_{p,i}=z_i+W_\xi\xi_{p,i},\\
\widehat K^{\mathrm{pre}}_{p,i}&=K^B_{p,i}+\widetilde z_{p,i}W^K_p+\zeta^K_p,\qquad
\widehat V_{p,i}=V^B_{p,i}+\widetilde z_{p,i}W^V_p+\zeta^V_p.
\end{aligned}
\label{app:eq:A6}
\end{equation}

The matrix $W_\xi$ projects the four conditions; $W_p^K,W_p^V\in\mathbb R^{w_W\times d_T}$ and $\zeta_p^K,\zeta_p^V$ are unit-specific readout biases. Conditioning and readout weights and biases are zero-initialized, so the initial output equals the Bridge. The Compiler does not separately read query hidden states; the query affects its output through selection.

Before mounting, apply target RoPE at original positions, $\widehat K_{p,i}=\operatorname{RoPE}_T(\widehat K^{\mathrm{pre}}_{p,i},i)$, retaining positions and validity masks. Query positions start at $N$ and generation continues at $N+Q$. External context and visible native query/generated KV share the target attention normalizer. Mandatory-tail states are also compiled, without additional native target context computation.

\FloatBarrier
\section{Supervision and Training Objectives}

\subsection{Local deletion utility and ranking}

Fix Q/K/V for a target attention head. Let $a_{j,i}$ be the weight on position $i$ in the visible history, $v_i$ its value, and $o_j=\sum_i a_{j,i}v_i$. Removing position $i$ and renormalizing gives the exact local identity:

\begin{equation}
o_j^{\setminus i}=\frac{o_j-a_{j,i}v_i}{1-a_{j,i}},\qquad
o_j^{\setminus i}-o_j=\frac{a_{j,i}}{1-a_{j,i}}(o_j-v_i).
\label{app:eq:B1}
\end{equation}

\begin{equation}
u_{p,i}=\frac1{|\mathcal O_p|}\sum_{j\in\mathcal O_p}
\left(\frac{a_{j,i}}{1-a_{j,i}}\right)^2
\frac{\|v_i-o_j\|_2^2}{\|o_j\|_2^2+\varepsilon_o}.
\label{app:eq:B2}
\end{equation}

The set $\mathcal O_p$ collects associated GQA query heads at up to eight final query positions and eight greedy teacher-continuation positions, sampling at most four target layers per example. The implementation floors $1-a_{j,i}$ and scales the stabilizer $\varepsilon_o$ using output energy. This utility measures a local change with fixed inputs, rather than the task loss from rerunning the entire model after deletion. Teacher continuation is used only for supervision.

\begin{equation}
\mathcal L_M=\operatorname{Mean}_{p}\operatorname{Mean}_{(i,j)\in\mathcal P_p}
\operatorname{softplus}\!\left(\frac{S_{p,j}-S_{p,i}}
{\max(\operatorname{Std}(S_{p,0:n}),10^{-4})}\right).
\label{app:eq:B3}
\end{equation}

Pairs satisfy $u_{p,i}>u_{p,j}$. Each valid head supplies 32 pairs: half straddle the uniform-budget utility cutoff and half come from local bands on either side. Ties in utility are excluded. Score-standard-deviation normalization limits improvements obtained merely by increasing score magnitude.

\subsection{Allocation objective and teacher definitions}

After freezing the Cartographer, the Allocator uses selected native target KV and minimizes teacher-forced answer NLL. It uses the first canonical answer, capped at 64 tokens; answers enter target supervision but not source-side ranking inputs. The extra output-KL weight is zero. The Compiler instead uses a same-selection native-KV teacher, extracting $\mathcal I$ from native target caches and recomputing the query:

\begin{equation}
p_T^{\mathcal I}(\cdot\mid q,c)=p_T(\cdot\mid q;\mathcal C_T(c)|_{\mathcal I}),
\qquad p_\psi^{\mathcal I}(\cdot\mid q,c)=p_T(\cdot\mid q;\widehat{\mathcal C}_{\psi}|_{\mathcal I}).
\label{app:eq:B4}
\end{equation}

The Compiler aligns these distributions to learn representation conversion at fixed positions and capacity. Both differ from the evaluation reference Target Full, which uses the entire original context. Selection and synthesis are separated in training design; this does not imply additive errors in downstream task metrics.

\subsection{Compiler geometry, attention, and behavior losses}

Let $K^*,V^*$ denote native target pre-RoPE keys and values at selected positions. Geometry terms average valid slots within each example before averaging examples. Define $r(v)=\sqrt{\|v\|_2^2/d_T+\epsilon}$ and $\nu(v)=\max(\|v\|_2,\epsilon)$, with $\epsilon=10^{-6}$.

\begin{equation}
\begin{aligned}
\mathcal L_{K,\mathrm{norm}}&=\operatorname{Mean}_{p,i}\frac1{d_T}
\left\|\frac{\widehat K^{\mathrm{pre}}_{p,i}}{r(\widehat K^{\mathrm{pre}}_{p,i})}
-\frac{K^*_{p,i}}{r(K^*_{p,i})}\right\|_2^2,\\
\mathcal L_{K,\mathrm{scale}}&=\operatorname{Mean}_{p,i}
[\log r(\widehat K^{\mathrm{pre}}_{p,i})-\log r(K^*_{p,i})]^2,\\
\mathcal L_V&=\operatorname{Mean}_{p,i}\left[
\left\|\frac{\widehat V_{p,i}}{\nu(\widehat V_{p,i})}-\frac{V^*_{p,i}}{\nu(V^*_{p,i})}\right\|_2^2
+[\log\nu(\widehat V_{p,i})-\log\nu(V^*_{p,i})]^2\right].
\end{aligned}
\label{app:eq:B5}
\end{equation}

Attention supervision uses up to the final 32 query positions from the native full-prompt forward pass. Fix teacher query vectors $q_j^*$ and visible native query keys $K_{q,\le j}^*$, replacing only selected context keys:

\begin{equation}
\begin{aligned}
a_j^*&=\operatorname{softmax}\!\left(q_j^*[K_{\mathcal I}^{*,\mathrm{rot}};K_{q,\le j}^*]^\top/\sqrt{d_T}\right),\\
\widehat a_j&=\operatorname{softmax}\!\left(q_j^*[\widehat K_{\mathcal I};K_{q,\le j}^*]^\top/\sqrt{d_T}\right),\qquad
\mathcal L_{K,\mathrm{attn}}=\operatorname{Mean}_{\ell,h_q,j}D_{\mathrm{KL}}(a_j^*\|\widehat a_j).
\end{aligned}
\label{app:eq:B6}
\end{equation}

Here $h_q$ indexes query heads, $\mathcal I$ is the selection of the corresponding physical KV head, and rot denotes target RoPE. Joint normalization preserves competition between context and query history. Following distribution-based knowledge distillation~\citep{hinton2015distillation}, behavioral supervision mounts native and compiled selections separately, recomputes the query, and compares full-vocabulary distributions at up to eight final query positions:

\begin{equation}
\mathcal L_{\mathrm{beh}}=\operatorname{Mean}_{j\in\mathcal J}
D_{\mathrm{KL}}(p_{T,j}^{\mathcal I}\|p_{\psi,j}^{\mathcal I}),\qquad |\mathcal J|=\min(8,Q).
\label{app:eq:B7}
\end{equation}

This KL uses temperature one and a stopped-gradient teacher; the compiled path retains input gradients. It uses no answer annotations and does not require generating eight answer tokens. The final objective is the five-term sum in the main compilation-supervision equation, with weight $10^{-4}$ only on key scale. Joint residual training begins from the Bridge plus zero residual, without separate key/value pretraining.

\FloatBarrier
\section{Budget Conservation and Gradient Derivations}
\label{app:budget-derivations}

\subsection{Continuous allocation, rounding, and uniqueness}

Let $t=\min(\tau,N)$ be the mandatory tail length, $k=\min(N,\max(t,\lceil\rho N\rceil))$ the average total capacity per head, $n=N-t$ the ranked-prefix length, and $B=P(k-t)$ the prefix budget. When $N\le\tau$, this gives $t=N$, so the prefix is empty and all context tokens are retained as the mandatory tail. For $0<B<Pn$, static learnable preferences $\alpha_p$ and a shared adjustment $\lambda$ give capacities:

\begin{equation}
b_p=n\sigma(\alpha_p-\lambda),\qquad \sum_{p=1}^P b_p=B.
\label{app:eq:C1}
\end{equation}

For finite $\alpha_p$, total capacity is continuous and strictly decreasing in $\lambda$, with limits $Pn$ and zero. Thus each interior budget has a unique solution, obtained by bisection. Boundaries directly return zero or full capacity. The mapping is also the first-order condition of the strictly concave problem below, where $\mathcal H(z)=-z\log z-(1-z)\log(1-z)$:

\begin{equation}
\max_{0\le b_p\le n,\ \sum_p b_p=B}\ \sum_p[\alpha_pb_p+n\mathcal H(b_p/n)].
\label{app:eq:C2}
\end{equation}

This optimization characterizes the budget map, not an optimum of the downstream task loss. Rounding starts with $f_p=\lfloor b_p\rfloor$ and assigns one extra slot to the $r=B-\sum_pf_p$ non-full heads with largest fractional parts, breaking ties deterministically. Since $B$ is integral, the resulting $k_p$ satisfy $\sum_pk_p=B$; adding tails gives $\sum_p|\mathcal I_p|=Pk$ without losing capacity through rounding.

\subsection{Derivatives of constrained allocation and soft rank gates}

Define $\omega_p=n\sigma(\alpha_p-\lambda)[1-\sigma(\alpha_p-\lambda)]$. Differentiation gives $\mathrm db_p=\omega_p(\mathrm d\alpha_p-\mathrm d\lambda)$. Conservation of the total budget yields

\begin{equation}
\mathrm d\lambda=\frac{\sum_p\omega_p\,\mathrm d\alpha_p}{\sum_p\omega_p},\qquad
\frac{\partial b_p}{\partial\alpha_j}=\omega_p\mathbf1[p=j]-\frac{\omega_p\omega_j}{\sum_r\omega_r}.
\label{app:eq:C3}
\end{equation}

Every first-order capacity change therefore sums to zero. For upstream gradient $v_p=\partial\mathcal L/\partial b_p$, backpropagation gives $\partial\mathcal L/\partial\alpha_p=\omega_p[v_p-\sum_jv_j\omega_j/\sum_j\omega_j]$, with zero gradients at boundaries. With frozen ranking and rank $r_{p,i}$, define soft gates whose total is $b_p$:

\begin{equation}
g_{p,i}=\sigma((\eta_p-r_{p,i})/T_R),\quad \sum_i g_{p,i}=b_p,
\qquad \frac{\partial g_{p,i}}{\partial b_p}=
\frac{g_{p,i}(1-g_{p,i})}{\sum_jg_{p,j}(1-g_{p,j})}.
\label{app:eq:C4}
\end{equation}

Set $T_R=2$ and solve $\eta_p$ by bisection; boundary capacities give all-zero or all-one gates. Implicit differentiation cancels the explicit temperature factor, while temperature still changes gates and the gradient distribution. The forward pass uses hard selection at integer capacities. Backpropagation uses a soft path that adds $\log g_{p,i}$ to context attention logits, retaining native query-history normalization. This is a biased surrogate for discrete selection, not the exact derivative of top-$k$.

\FloatBarrier
\section{Training Configuration and Data Boundaries}
\label{app:training-configuration}

We train a separate interface for each direction, freezing both base models throughout. Main runs use $\rho=0.25$ and $\tau=32$. The Cartographer and Allocator use a fixed pool of 5,000 SearchQA training examples; the Compiler uses a fixed 500-example subset. The Cartographer additionally uses 1,000 validation examples for generation evaluation at the end of its stage.

\begin{table}[htbp]
\centering\footnotesize
\setlength{\tabcolsep}{4pt}
\caption{Recorded training schedules.}
\begin{tabular}{lrrr}
\toprule
Setting & Cartographer & Allocator & Compiler residual \\
\midrule
Training examples & 5,000 & 5,000 & 500 \\
Epochs & 1 & 1 & 4 \\
Global batch & 16 & 4 & 4 \\
Optimizer updates & 313 & 1,250 & 500 \\
Optimizer & AdamW & AdamW & AdamW \\
Learning rate & $10^{-4}$ & $10^{-2}$ & $10^{-4}$ \\
Weight decay & 0.01 & 0 & 0.01 \\
Gradient clip & 1 & 1 & 1 \\
Seed & 17 & 434781 & 469601542 \\
\bottomrule
\end{tabular}
\end{table}

All three trainable stages use AdamW~\citep{loshchilov2019adamw}. The Cartographer's final batch contains eight examples. Four Compiler epochs correspond to 2,000 example presentations and 500 updates; Bridge fitting makes an additional pass over 500 examples. The Allocator and Compiler retain their fixed training endpoints without an additional best-epoch search; the Cartographer uses stage-end validation generation metrics. Data and optimization costs are reported by stage: the complete pipeline is not trained on only 500 examples.

Source forwards use BF16; Cartographer ranking and its teacher use FP32. The Allocator and Compiler use the frozen target's BF16 path. Compiler forwards use BF16 autocast, with critical normalizations and losses in FP32. Freezing target parameters does not remove gradients to interface inputs. We freeze $\theta$ after Cartographer training, $\alpha$ after Allocator training, and $\beta$ after Bridge fitting, then optimize only $\psi$.

Compression settings are trained separately, keeping Compiler width, depth, and training schedule fixed within each direction. The $1\times$ setting bypasses ranking and allocation but still learns compilation, so it is not Target Full. 

\FloatBarrier
\section{Pair Capacity Rules and Parameter Counts}

Model sources are the Qwen3~\citep{yang2025qwen3}, Qwen2.5~\citep{yang2024qwen25}, and Llama 3~\citep{grattafiori2024llama3} technical reports and the official Llama 3.2 model card~\citep{meta2024llama32}.

\subsection{Capacity from model geometry}
\label{app:capacity-geometry}

Source hidden width $D_S$ determines input projections; $P=L_TH_T$ and target head dimension $d_T$ determine output geometry. To adapt the shared trunk as well, let $D_T$ be target hidden width and define the following preset rule, where $\lceil x\rceil_a=a\lceil x/a\rceil$ denotes upward alignment:

\begin{equation}
\begin{aligned}
s&=\max\{1,\sqrt{D_S/1024},\sqrt{D_T/4096},\sqrt{P/288}\},\\
(w_M,d_M,c_M)&=(\lceil256s\rceil_{32},\lceil64s\rceil_{16},\lceil32s\rceil_8),\qquad m_M=4,\\
w_W&=\lceil256s\rceil_{32},\quad m_W=\lceil16s\rceil_4,\quad
D_Z=D_U=\max(2,\lceil2s\rceil),\quad f_W=\lceil2w_W\rceil_8.
\end{aligned}
\label{app:eq:E1}
\end{equation}

Compiler attention uses dimension 32 per head, hence $w_W/32$ heads, and retains three intermediate taps. The Allocator always contains only $P$ parameters. This rule standardizes configurations rather than asserting an empirically optimal scaling law. Source width, target geometry, and shared capacity can all change in the reverse direction, whose interface is trained separately.

\paragraph{Limitations under strong size asymmetry.}
The current rule sets interface capacity from source and target hidden widths
and target KV geometry, without explicitly accounting for the directional
model-size ratio, representation mismatch, or their training requirements.
When the source is much larger than the target, a high source-to-target size
ratio may lead to degradation under the current configuration; this ratio
refers to model parameter counts, not KV compression.
The Qwen3 $14\mathrm{B}\!\to\!4\mathrm{B}$ results in
Appendix~\ref{app:additional-pairs} show that the current capacity rule does not
ensure contextual-task transfer quality. Our cross-model parameterization therefore remains
incomplete. Future work will investigate how asymmetric source and target
scales affect context reuse, and how interface capacity, depth, and training
budget should be adapted jointly.

\subsection{Exact counts and validation limits}

\begin{equation}
\begin{aligned}
P_W={}&(4D_S+1)w_W+4w_W+m_Ww_W\\
&+D_Z(8w_W^2+3w_Wf_W+3w_W)\\
&+D_U(4w_W^2+3w_Wf_W+2w_W)+w_W+2P(w_Wd_T+d_T),\\
P_M={}&D_Sw_M+2D_S+7w_M+4w_Md_M\\
&+(L_T+H_T)c_M+2c_Md_M+5d_M^2+15d_M,\\
P_R={}&P,\qquad P_B=2Pd_T(d_S+1).
\end{aligned}
\label{app:eq:E2}
\end{equation}

Here $P_W$ counts only the Compiler residual network and $P_B$ its internal Bridge. The complete Compiler has $P_W+P_B$ parameters; total interface size is $P_M+P_R+P_W+P_B$, of which $P_M+P_R+P_W$ are optimized by gradients. Base-model weights are excluded. Unit-specific readouts already scale with target geometry and receive no additional model-size multiplier.

\begin{equation}
C_M=\left\lceil10^6\max(s^2,P/288)\right\rceil,\qquad
C_W=10^6\left\lceil1.25P_W/10^6\right\rceil.
\label{app:eq:E3}
\end{equation}

The limits $C_M,C_W$ validate instantiated parameter counts; they introduce no parameters and are not target sizes. The limit $C_W$ applies to the residual network; the complete Compiler's corresponding limit is $C_W+P_B$. Command-line overrides of widths, depths, or slot counts must be saved together with the resulting validation.

\begin{table}[htbp]
\centering\footnotesize
\setlength{\tabcolsep}{4pt}
\caption{Recorded geometry for six directions; D is the shared encoder/decoder depth.}
\begin{tabular}{lrrrr}
\toprule
Pair & $D_S$ & $(L_T,H_T)$ & $(w_M,d_M,c_M)$ & $(w_W,m_W,D)$ \\
\midrule
Qwen3 4B$\to$8B & 2560 & (36,8) & (416,112,56) & (416,28,4) \\
Qwen3 8B$\to$4B & 4096 & (36,8) & (512,128,64) & (512,32,4) \\
Llama 3B$\to$8B & 3072 & (32,8) & (448,112,56) & (448,28,4) \\
Llama 8B$\to$3B & 4096 & (28,8) & (512,128,64) & (512,32,4) \\
Qwen2.5 7B$\to$32B & 3584 & (64,8) & (480,128,64) & (480,32,4) \\
Qwen2.5 32B$\to$7B & 5120 & (28,4) & (576,144,72) & (576,36,5) \\
\bottomrule
\end{tabular}
\end{table}

\begin{table}[htbp]
\centering\footnotesize
\setlength{\tabcolsep}{4pt}
\caption{Exact interface parameter counts; Compiler includes the fitted, frozen Bridge.}
\begin{tabular}{lrrrr}
\toprule
Pair & Cartographer & Allocator & Compiler & Total \\
\midrule
Qwen3 4B$\to$8B & 1,338,768 & 288 & 61,151,168 & 62,490,224 \\
Qwen3 8B$\to$4B & 2,474,112 & 288 & 80,917,504 & 83,391,904 \\
Llama 3B$\to$8B & 1,665,424 & 256 & 62,676,608 & 64,342,288 \\
Llama 8B$\to$3B & 2,473,600 & 224 & 70,398,976 & 72,872,800 \\
Qwen2.5 7B$\to$32B & 2,081,440 & 512 & 108,981,440 & 111,063,392 \\
Qwen2.5 32B$\to$7B & 3,424,048 & 112 & 71,890,624 & 75,314,784 \\
\bottomrule
\end{tabular}
\end{table}

\begin{table}[htbp]
\centering\footnotesize
\setlength{\tabcolsep}{4pt}
\caption{Compiler decomposition and gradient-optimized parameters of the entire interface.}
\begin{tabular}{lrrr}
\toprule
Pair & Bridge & Residual & Gradient-optimized total \\
\midrule
Qwen3 4B$\to$8B & 9,510,912 & 51,640,256 & 52,979,312 \\
Qwen3 8B$\to$4B & 9,510,912 & 71,406,592 & 73,880,992 \\
Llama 3B$\to$8B & 8,454,144 & 54,222,464 & 55,888,144 \\
Llama 8B$\to$3B & 7,397,376 & 63,001,600 & 65,475,424 \\
Qwen2.5 7B$\to$32B & 16,908,288 & 92,073,152 & 94,155,104 \\
Qwen2.5 32B$\to$7B & 3,698,688 & 68,191,936 & 71,616,096 \\
\bottomrule
\end{tabular}
\end{table}

Counts are derived from the actual dimensions in six recorded configurations; residual counts match instantiation preflights. All six directions use $d_S=d_T=128$. The general rectangular-map count does not establish execution support for every unequal-head-dimension pair. The interface is lightweight relative to billion-parameter bases; latency also depends on activations, readouts, and mounting.

\FloatBarrier
\section{Execution Interface, Complexity, and Reuse}
\label{app:execution-interface}

\subsection{Token and position interfaces}

We verify token-to-ID semantics within each Qwen3, Llama, and Qwen2.5 pair and use the target tokenizer consistently. Llama tokenizer JSON files need not have identical file hashes, but their actual vocabulary mappings must agree. In-range IDs alone do not establish semantic compatibility. Context, query, and generation boundaries are matched across training, native Full, and mounted execution.

We retain native chat templates where available. Llama Base contexts contain a single BOS, context text, and two newlines. Contextual QA uses completion-v2, appending the continuation cue ``Answer: The answer is'' to the query, or only ``The answer is'' after an existing ``Answer:''. General choice and language-modeling tasks retain plain-v1 and stored token sequences. This dataset-specific protocol is matched between the method and Full.

Mounting native target KV at every position controls interface consistency. We test a representative example per dataset plus the longest general-task example, comparing up to four generated tokens and EOS behavior; candidate-token NLL differences must not exceed 0.05 in absolute value. This checks cache, position, and attention implementation, without requiring compiled KV to equal native KV.

\subsection{Effective state size and computation}

\begin{equation}
\operatorname{Elements}(\widehat{\mathcal C})=2d_T\sum_p|\mathcal I_p|=2Pkd_T,
\qquad \frac{\operatorname{Elements}(\widehat{\mathcal C})}{2PNd_T}=\frac{k}{N}.
\label{app:eq:F1}
\end{equation}

For $N=1024$, $\rho=0.25$, $\tau=32$, and $P=288$, there are 73,728 effective slots versus 294,912 in the full cache, with the tail included. For union size $U=|\bigcup_p\mathcal I_p|$, we have $\max_p|\mathcal I_p|\le U\le\min(N,Pk)$. Retaining one quarter of the total KV slots does not imply that the union of retained positions covers only one quarter of the context.

Cartographer scores have size $O(PN)$; fixed query slots do not construct an $N\times N$ target attention matrix. Let $S_{\max}=\max_p|\mathcal I_p|$. The main residual-network cost is

\begin{equation}
\begin{aligned}
\mathcal O\big(&ND_Sw_W+D_Z[(N+m_W)w_W^2+m_WNw_W+m_W^2w_W]\\
&+D_U[(U+m_W)w_W^2+Um_Ww_W]+PS_{\max}w_Wd_T\big).
\end{aligned}
\label{app:eq:F2}
\end{equation}

The shared context and union trunk carry no additional factor $P$, while per-unit readouts depend on target geometry. Current readouts pad heads to a common width, hence $PS_{\max}$. Bridge affine maps add their own per-slot cost.

Latent slots read the complete provided context and can aggregate context positions later than a given slot. Those positions are available for queries after the context, as evaluated here.

\subsection{Complete and source-state reuse paths}

\begin{equation}
\begin{aligned}
T_{\mathrm{ACCT}}^{\mathrm{prefill}}&=T_S([c;q])+T_{\mathrm{transfer}}(c,q)
+T_{T,q}(q;\widehat{\mathcal C}),\\
T_{\mathrm{ACCT}}^{\mathrm{incremental}}&=T_{S,q}(q;\mathcal R_S(c))
+T_{\mathrm{transfer}}(c,q)+T_{T,q}(q;\widehat{\mathcal C}).
\end{aligned}
\label{app:eq:F3}
\end{equation}

The first line is the main paper's complete-cost decomposition. Transfer includes the three components, sorting, mounting, and required data movement. The second line assumes that $\mathcal R_S(c)=(H_c,H_S,K_S,V_S)$ already exists and still includes source query computation.

When the query changes, causal source context states can be reused if weights, tokens, and positions are unchanged, but ranking, selection, and compilation must be repeated. Allocator counts remain fixed at the same length and budget while content can change.

\FloatBarrier
\section{Data and Evaluation Metrics}
\label{app:data-evaluation}

\begin{table}[htbp]
\centering\footnotesize
\setlength{\tabcolsep}{4pt}
\caption{The fixed 10,000-example evaluation panel.}
\begin{tabular}{llrl}
\toprule
Group & Dataset & Examples & Metric \\
\midrule
General & HellaSwag~\citep{zellers-etal-2019-hellaswag} & 1,000 & Accuracy \\
General & Winogrande~\citep{sakaguchi2020winogrande} & 1,000 & Accuracy \\
General & MMLU~\citep{hendrycks2021mmlu} & 1,000 & Accuracy \\
General & PIQA~\citep{bisk2020piqa} & 1,000 & Accuracy \\
General & WikiText2~\citep{merity2017wikitext} & 500 & PPL \\
Context & SearchQA~\citep{dunn2017searchqa} & 1,000 & Token F1 \\
Context & TriviaQA~\citep{joshi-etal-2017-triviaqa} & 1,000 & Token F1 \\
Context & BoolQ~\citep{clark-etal-2019-boolq} & 1,000 & Accuracy \\
Context & RACE~\citep{lai-etal-2017-race} & 1,000 & Accuracy \\
Context & WikiHop~\citep{welbl-etal-2018-constructing} & 1,000 & Accuracy \\
Context & PubMedQA~\citep{jin-etal-2019-pubmedqa} & 500 & Accuracy \\
\bottomrule
\end{tabular}
\end{table}

SearchQA and TriviaQA use token F1; all other tasks except WikiText2 use accuracy. WikiText2 PPL is the exponential of mean NLL over evaluated tokens, not a mean of example-level perplexities. General multiple-choice tasks use candidate-sequence NLL, preserving exact token IDs, options, and budgets in stored rows. Contextual tasks primarily score generated answers.

Generation-based evaluations use greedy decoding with a 32,768-token input limit, 32 reserved tokens, and at most 32 generated tokens, following shared EOS handling and answer normalization. All six directions use the same fixed examples. GSM8K is excluded.

Target Full results are reused only when target-weight identity, actual input tokens, positions, precision, prompts, candidates, decoding budgets, and scoring code match. A source-model change alone does not require rerunning an identical Full contract, while method outputs and interface training remain specific to each run.

\FloatBarrier
\section{Additional Analysis of Recorded Results}

The following table is computed from the updated main-table entries. Retention is ACCT/Full for higher-is-better metrics and Full/ACCT for PPL. We average equally across five general benchmarks and separately across six contextual tasks. 

\begin{table}[htbp]
\centering\footnotesize
\setlength{\tabcolsep}{4pt}
\caption{Descriptive retention and task wins derived from the recorded main-table entries.}
\begin{tabular}{lrrr}
\toprule
Pair & General retention & Context retention & Wins over Full \\
\midrule
Qwen3 4B$\to$8B & 101.7\% & 95.9\% & 4/11 \\
Qwen3 8B$\to$4B & 94.7\% & 106.6\% & 4/11 \\
Llama 3B$\to$8B & 64.7\% & 77.5\% & 2/11 \\
Llama 8B$\to$3B & 67.2\% & 74.7\% & 2/11 \\
Qwen2.5 7B$\to$32B & 66.8\% & 77.5\% & 1/11 \\
Qwen2.5 32B$\to$7B & 75.7\% & 63.4\% & 0/11 \\
\bottomrule
\end{tabular}
\end{table}

Both Qwen3 directions improve open-domain QA, while Llama and Qwen2.5 have substantial gaps on some language-modeling, reading-comprehension, and multi-hop tasks. The results support directed-interface execution and task-specific transfer gains. In particular, original-scale PPL remains necessary so inverse-ratio summaries do not hide large losses.

In the compression study, increasing capacity from $4\times$ to $2\times$ improves all six contextual tasks in the reverse direction, while full retention is not uniformly best. The $1\times$ configuration also changes whether selection is active, so it compares complete budget configurations.

\section{Per-Dataset Results for Additional Model Pairs}
\label{app:additional-pairs}

Table~\ref{tab:additional-pairs} supplements the main table with seven Qwen3
directions: $0.6\mathrm{B}\!\to\!4\mathrm{B}$,
$0.6\mathrm{B}\!\to\!8\mathrm{B}$, $0.6\mathrm{B}\!\to\!14\mathrm{B}$,
$4\mathrm{B}\!\leftrightarrow\!14\mathrm{B}$, and
$8\mathrm{B}\!\leftrightarrow\!14\mathrm{B}$.
These runs use the mainline configuration of $4\times$ compression and 500
Compiler training examples for four epochs, with the same 10,000 evaluation
examples and main-table metrics. All entries come from completed evaluations
with zero execution failures; Target Full is the native target reference for
the corresponding run.

\begin{table}[htbp]
\centering\footnotesize
\setlength{\tabcolsep}{1.2pt}
\renewcommand{\arraystretch}{1.1}
\caption{Additional Qwen3 results on 11 datasets (F1/accuracy in \%; PPL $\downarrow$), with bold marking ACCT improvements over the corresponding Target Full.}
\label{tab:additional-pairs}
\begin{tabular*}{\textwidth}{@{\extracolsep{\fill}}ll*{11}{r}@{}}
\toprule
& & \multicolumn{5}{c}{General capability} & \multicolumn{6}{c}{Context-dependent tasks} \\
\cmidrule(lr){3-7}\cmidrule(lr){8-13}
Model pair & Method & \shortstack{Hella\\Swag} & \shortstack{Wino\\grande} & MMLU & PIQA & \shortstack{Wiki\\Text2} & \shortstack{Search\\QA} & \shortstack{Trivia\\QA} & BoolQ & RACE & WikiHop & \shortstack{PubMed\\QA} \\
 & & Acc. & Acc. & Acc. & Acc. & PPL $\downarrow$ & F1 & F1 & Acc. & Acc. & Acc. & Acc. \\
\midrule
\multirow{2}{*}{\shortstack[l]{Qwen3\\$0.6\mathrm{B}\!\to\!4\mathrm{B}$}} & Target Full & 37.10 & 51.00 & 63.60 & 69.20 & 81.29 & 38.56 & 37.68 & 84.00 & 87.10 & 45.80 & 34.00 \\
 & ACCT & 34.60 & 49.70 & 60.70 & 65.40 & \textbf{46.66} & \textbf{44.29} & \textbf{45.63} & 68.10 & 53.20 & 32.60 & \textbf{49.00} \\
\midrule
\multirow{2}{*}{\shortstack[l]{Qwen3\\$0.6\mathrm{B}\!\to\!8\mathrm{B}$}} & Target Full & 39.70 & 58.50 & 59.70 & 73.20 & 32.91 & 46.65 & 50.48 & 86.30 & 87.60 & 49.60 & 48.80 \\
 & ACCT & 33.50 & 50.40 & 59.00 & 65.30 & 38.10 & 40.29 & 41.70 & 62.80 & 53.90 & 28.20 & 43.40 \\
\midrule
\multirow{2}{*}{\shortstack[l]{Qwen3\\$0.6\mathrm{B}\!\to\!14\mathrm{B}$}} & Target Full & 36.60 & 59.20 & 72.30 & 73.90 & 19.74 & 65.45 & 52.83 & 88.60 & 91.10 & 49.10 & 47.40 \\
 & ACCT & 36.30 & 52.00 & 68.20 & 69.80 & 21.37 & 57.43 & \textbf{56.44} & 64.00 & 60.20 & 36.50 & 19.00 \\
\midrule
\multirow{2}{*}{\shortstack[l]{Qwen3\\$4\mathrm{B}\!\to\!14\mathrm{B}$}} & Target Full & 36.60 & 59.20 & 72.30 & 73.90 & 19.74 & 65.45 & 52.83 & 88.60 & 91.10 & 49.10 & 47.40 \\
 & ACCT & \textbf{36.90} & 49.30 & 52.40 & 68.20 & 24.02 & 53.91 & \textbf{57.13} & 67.10 & 59.10 & 35.10 & 39.80 \\
\midrule
\multirow{2}{*}{\shortstack[l]{Qwen3\\$14\mathrm{B}\!\to\!4\mathrm{B}$}} & Target Full & 37.10 & 51.00 & 63.60 & 69.20 & 81.29 & 38.56 & 37.68 & 84.00 & 87.10 & 45.80 & 34.00 \\
 & ACCT & 25.60 & 50.40 & 25.50 & 55.40 & 98.38 & 0.31 & 0.66 & 0.60 & 13.80 & 0.60 & 0.60 \\
\midrule
\multirow{2}{*}{\shortstack[l]{Qwen3\\$8\mathrm{B}\!\to\!14\mathrm{B}$}} & Target Full & 36.60 & 59.20 & 72.30 & 73.90 & 19.74 & 65.45 & 52.83 & 88.60 & 91.10 & 49.10 & 47.40 \\
 & ACCT & \textbf{37.40} & 49.50 & 69.00 & 70.40 & \textbf{19.03} & 58.93 & \textbf{61.59} & 69.10 & 59.00 & 37.40 & 39.80 \\
\midrule
\multirow{2}{*}{\shortstack[l]{Qwen3\\$14\mathrm{B}\!\to\!8\mathrm{B}$}} & Target Full & 39.70 & 58.50 & 59.70 & 73.20 & 32.91 & 46.65 & 50.48 & 86.30 & 87.60 & 49.60 & 48.80 \\
 & ACCT & 30.60 & 50.90 & 55.80 & 65.10 & 53.96 & 20.20 & 26.94 & 48.90 & 49.50 & 26.90 & \textbf{54.00} \\
\bottomrule
\end{tabular*}
\end{table}

\paragraph{Performance across directions.}
The $0.6\mathrm{B}\!\to\!4\mathrm{B}$ configuration exceeds Target Full on both
open-domain QA tasks, and all three additional directions targeting 14B improve
TriviaQA. Qwen3 $14\mathrm{B}\!\to\!4\mathrm{B}$ remains below 1\% accuracy
on several contextual tasks.
Its contextual-task degradation should therefore be considered separately from
language modeling, and transfer quality remains dependent on the pair and task.

\FloatBarrier
\section{Component Ablation Protocols}
\label{app:compiler-ablation}

All controls use the formal Qwen3 $4\mathrm{B}\!\leftrightarrow\!8\mathrm{B}$
configurations with a $4\times$ budget and the mandatory 32-token tail.
Random deterministically generates independent prefix rankings for each
layer/head from a fixed seed and the exact prompt tokens; the same rule is used
in training and evaluation, without reference answers. We retrain the Allocator
on the original 5,000 SearchQA examples, then refit the Bridge and train the
Compiler. Uniform retains the original Cartographer, assigns identical counts
to all layer/head units, and refits the Bridge and retrains the Compiler.
Both use the original 500-example subset, four epochs, 500 updates, and
direction-specific Compiler dimensions.

Bridge-only and Native KV fix the original Cartographer, Allocator, and
retained layer/head positions without retraining. Bridge-only retains the
fitted affine Bridge and removes only the neural residual. Native KV gathers
target-native context KV at identical positions, preserving existing key RoPE
without a second rotation; it still requires target context prefill. These
controls assess residual correction and state quality at fixed selection,
rather than retraining after removing the entire Compiler.

All eight runs complete 10,000 examples with zero execution failures or duplicate
samples. Across batches, we verify sample identities, exact input tokens,
context/query boundaries, total retained slots, and reused per-example ACCT/Full
baselines; the two Compiler controls also match exact selected positions.
Each method uses a fresh target continuation cache. F1 and accuracy are averaged
over examples; WikiText2 PPL is the exponential of total token NLL divided by
total scored tokens.
\FloatBarrier

\section{Inference Latency Protocol and Results by Context Length}
\label{app:latency-protocol}

\paragraph{Hardware and concurrency.}
We launch six independent workers on a machine with eight NVIDIA A800 80GB
GPUs, with one process per transfer direction. The four Qwen3
$4\mathrm{B}\!\leftrightarrow\!8\mathrm{B}$ and Llama
$3\mathrm{B}\!\leftrightarrow\!8\mathrm{B}$ directions each occupy one of
GPUs 0--3, placing the source, target, and three components on the same device.
Each Qwen2.5 $7\mathrm{B}\!\leftrightarrow\!32\mathrm{B}$ direction uses two
GPUs, in groups 4--5 and 6--7. Within each process, the source and components
reside on its first visible GPU and the target on its second. This is
concurrency across directions; individual directions use neither tensor nor
data parallelism. We use PyTorch~\citep{paszke2019pytorch} with BF16 and SDPA, disable \texttt{torch.compile}
(\texttt{SLOT\_COMPILE=0}), and set OMP, MKL, and OpenBLAS CPU thread counts to 4.

\paragraph{Samples and timing.}
Each fixed sample runs independently with batch size 1 and input shape $[1,L]$.
Measurements use $\rho=0.25$ and a 32-token tail.
We measure eight context samples per direction; their actual lengths are
listed below. Source prefill, Cartographer, Allocator, Compiler, and target
prefill are each warmed up and then timed according to
Table~\ref{tab:latency-repetitions}, reporting the median. Before timing,
\texttt{cuda.synchronize} synchronizes all GPUs visible to the process;
\texttt{time.perf\_counter()} records wall-clock time. Repetition counts apply
to each stage of each sample.

\begin{table}[htbp]
\centering\small
\caption{Warm-up and timed repetitions for each stage.}
\label{tab:latency-repetitions}
\begin{tabular}{lrr}
\toprule
Context length $L$ & Warm-up runs & Timed runs \\
\midrule
$L<4096$ & 2 & 3 \\
$4096\le L<8192$ & 1 & 3 \\
$L\ge8192$ & 1 & 2 \\
\bottomrule
\end{tabular}
\end{table}

\paragraph{Comparison scope and results.}
KV state-conversion latency is the sum of synchronized wall-clock medians for
Cartographer, Allocator, and Compiler. Speedup is target context prefill latency
divided by this sum. Source prefill is tabulated separately; Section~3 and
Appendix~F.3 give the complete prefill cost decomposition.
Tables~\ref{tab:latency-qwen3}--\ref{tab:latency-qwen25} report stage times and three-component totals in ms;
Cart., Alloc., and Comp. denote the Cartographer, Allocator, and Compiler.
Figure~\ref{fig:context-reuse-latency} uses all 48 measurements. Tabulated times
and speedups are rounded to one and two decimal places, respectively.

\clearpage
\begin{table}[!ht]
\centering\footnotesize
\setlength{\tabcolsep}{2.5pt}
\renewcommand{\arraystretch}{1.08}
\caption{Qwen3 stage latencies (ms) and state-conversion speedups by context length.}
\label{tab:latency-qwen3}
\begin{tabular*}{\textwidth}{@{\extracolsep{\fill}}l*{8}{r}@{}}
\toprule
Direction & $L$ & Cart. & Alloc. & Comp. & Convert & Source & Target & Speedup \\
\midrule
\multirow{8}{*}{$4\mathrm{B}\!\to\!8\mathrm{B}$} & 1001 & 7.0 & 7.4 & 12.8 & 27.2 & 69.4 & 91.1 & $3.35\times$ \\
 & 1024 & 7.0 & 7.4 & 12.9 & 27.3 & 69.4 & 91.9 & $3.37\times$ \\
 & 1729 & 7.0 & 7.9 & 17.0 & 31.9 & 106.1 & 159.0 & $4.98\times$ \\
 & 1918 & 7.1 & 7.4 & 18.1 & 32.6 & 131.6 & 165.3 & $5.07\times$ \\
 & 4126 & 15.6 & 7.5 & 30.6 & 53.7 & 270.2 & 363.9 & $6.78\times$ \\
 & 8158 & 20.3 & 7.8 & 53.3 & 81.4 & 533.7 & 729.8 & $8.97\times$ \\
 & 16389 & 43.5 & 7.8 & 105.0 & 156.3 & 1242.7 & 1658.2 & $10.61\times$ \\
 & 19070 & 48.0 & 7.7 & 124.6 & 180.3 & 1496.7 & 1993.4 & $11.06\times$ \\
\midrule
\multirow{8}{*}{$8\mathrm{B}\!\to\!4\mathrm{B}$} & 1001 & 7.0 & 7.4 & 13.0 & 27.4 & 102.5 & 69.3 & $2.53\times$ \\
 & 1024 & 6.9 & 7.3 & 12.9 & 27.1 & 103.2 & 57.5 & $2.12\times$ \\
 & 1729 & 7.0 & 7.3 & 17.2 & 31.5 & 170.9 & 98.4 & $3.12\times$ \\
 & 1918 & 7.0 & 7.3 & 18.3 & 32.6 & 200.8 & 107.6 & $3.30\times$ \\
 & 4126 & 15.5 & 7.4 & 31.3 & 54.2 & 412.8 & 242.6 & $4.48\times$ \\
 & 8158 & 20.4 & 7.4 & 55.9 & 83.7 & 791.0 & 492.6 & $5.89\times$ \\
 & 16389 & 43.0 & 8.6 & 110.5 & 162.1 & 1765.4 & 1188.5 & $7.33\times$ \\
 & 19070 & 46.4 & 8.2 & 128.1 & 182.7 & 2127.7 & 1456.3 & $7.97\times$ \\
\bottomrule
\end{tabular*}
\end{table}

\begin{table}[!ht]
\centering\footnotesize
\setlength{\tabcolsep}{2.5pt}
\renewcommand{\arraystretch}{1.08}
\caption{Llama stage latencies (ms) and state-conversion speedups by context length.}
\label{tab:latency-llama}
\begin{tabular*}{\textwidth}{@{\extracolsep{\fill}}l*{8}{r}@{}}
\toprule
Direction & $L$ & Source & Cart. & Alloc. & Comp. & Convert & Target & Speedup \\
\midrule
\multirow{8}{*}{$3\mathrm{B}\!\to\!8\mathrm{B}$} & 995 & 42.6 & 6.2 & 7.4 & 12.3 & 25.9 & 86.4 & $3.34\times$ \\
 & 996 & 43.8 & 6.2 & 7.4 & 12.3 & 25.9 & 86.4 & $3.34\times$ \\
 & 1551 & 64.5 & 6.2 & 7.4 & 15.3 & 28.9 & 123.8 & $4.28\times$ \\
 & 1900 & 88.2 & 6.2 & 7.4 & 17.1 & 30.7 & 153.1 & $4.99\times$ \\
 & 3989 & 176.3 & 9.9 & 7.4 & 28.3 & 45.6 & 319.9 & $7.02\times$ \\
 & 7848 & 339.0 & 17.6 & 7.4 & 49.4 & 74.4 & 656.8 & $8.83\times$ \\
 & 15947 & 792.9 & 33.4 & 7.7 & 98.3 & 139.4 & 1477.9 & $10.60\times$ \\
 & 19047 & 995.8 & 40.8 & 7.7 & 117.1 & 165.6 & 1837.5 & $11.10\times$ \\
\midrule
\multirow{8}{*}{$8\mathrm{B}\!\to\!3\mathrm{B}$} & 995 & 90.7 & 7.5 & 7.3 & 11.3 & 26.1 & 39.8 & $1.52\times$ \\
 & 996 & 93.8 & 5.5 & 7.5 & 11.1 & 24.1 & 45.7 & $1.90\times$ \\
 & 1551 & 129.4 & 5.6 & 7.0 & 13.5 & 26.1 & 60.7 & $2.33\times$ \\
 & 1900 & 178.0 & 5.6 & 7.0 & 15.1 & 27.7 & 73.3 & $2.65\times$ \\
 & 3989 & 351.8 & 8.8 & 6.9 & 24.3 & 40.0 & 156.4 & $3.91\times$ \\
 & 7848 & 681.6 & 15.7 & 6.9 & 42.4 & 65.0 & 318.2 & $4.90\times$ \\
 & 15947 & 1526.1 & 29.1 & 7.2 & 84.2 & 120.5 & 742.1 & $6.16\times$ \\
 & 19047 & 1905.4 & 36.0 & 7.4 & 100.4 & 143.8 & 937.4 & $6.52\times$ \\
\bottomrule
\end{tabular*}
\end{table}

\clearpage
\begin{table}[!ht]
\centering\footnotesize
\setlength{\tabcolsep}{2.5pt}
\renewcommand{\arraystretch}{1.08}
\caption{Qwen2.5 stage latencies (ms) and state-conversion speedups by context length.}
\label{tab:latency-qwen25}
\begin{tabular*}{\textwidth}{@{\extracolsep{\fill}}l*{8}{r}@{}}
\toprule
Direction & $L$ & Source & Cart. & Alloc. & Comp. & Convert & Target & Speedup \\
\midrule
\multirow{8}{*}{$7\mathrm{B}\!\to\!32\mathrm{B}$} & 1001 & 85.7 & 10.7 & 8.0 & 17.2 & 35.9 & 308.5 & $8.59\times$ \\
 & 1024 & 86.4 & 10.8 & 8.3 & 17.3 & 36.4 & 310.9 & $8.54\times$ \\
 & 1729 & 127.3 & 11.0 & 8.2 & 24.4 & 43.6 & 552.9 & $12.68\times$ \\
 & 1918 & 162.2 & 10.9 & 8.3 & 26.2 & 45.4 & 584.1 & $12.87\times$ \\
 & 4126 & 323.6 & 25.9 & 8.3 & 48.6 & 82.8 & 1324.1 & $15.99\times$ \\
 & 8158 & 625.1 & 34.0 & 8.4 & 89.1 & 131.5 & 2592.0 & $19.71\times$ \\
 & 16389 & 1348.4 & 73.8 & 8.9 & 193.1 & 275.8 & 5653.4 & $20.50\times$ \\
 & 19070 & 1607.3 & 83.3 & 9.0 & 223.7 & 316.0 & 6706.3 & $21.22\times$ \\
\midrule
\multirow{8}{*}{$32\mathrm{B}\!\to\!7\mathrm{B}$} & 1001 & 355.4 & 3.9 & 7.1 & 10.7 & 21.7 & 76.6 & $3.53\times$ \\
 & 1024 & 357.6 & 3.9 & 7.4 & 10.7 & 22.0 & 76.3 & $3.47\times$ \\
 & 1729 & 571.1 & 4.0 & 7.1 & 12.5 & 23.6 & 123.4 & $5.23\times$ \\
 & 1918 & 706.7 & 4.1 & 7.4 & 12.9 & 24.4 & 131.6 & $5.39\times$ \\
 & 4126 & 1413.2 & 7.6 & 7.2 & 18.2 & 33.0 & 303.4 & $9.19\times$ \\
 & 8158 & 2689.8 & 12.8 & 7.4 & 29.9 & 50.1 & 595.7 & $11.89\times$ \\
 & 16389 & 5664.2 & 24.7 & 7.5 & 53.7 & 85.9 & 1322.7 & $15.40\times$ \\
 & 19070 & 6732.5 & 28.2 & 7.5 & 61.6 & 97.3 & 1580.7 & $16.25\times$ \\
\bottomrule
\end{tabular*}
\end{table}

\FloatBarrier

\paragraph{Choosing a source when states are unavailable.}
When no source states are available and the goal is to construct target KV
states from raw context, source encoding must also enter the latency budget.
A speedup requires source prefill plus state transfer to be faster than direct
target prefill. A smaller source leaves more room for transfer overhead and is
therefore the preferred starting point for this use case.
Table~\ref{tab:latency-with-source} sums the measured stage medians for source
prefill, the three components, packing, and positional rotation; subsequent
target query encoding and generation are outside this state-construction metric.

At approximately 19K tokens, Qwen3 $4\mathrm{B}\!\to\!8\mathrm{B}$,
Llama $3\mathrm{B}\!\to\!8\mathrm{B}$, and Qwen2.5
$7\mathrm{B}\!\to\!32\mathrm{B}$ achieve $1.16\times$, $1.53\times$, and
$3.35\times$ speedups after accounting for source encoding, reducing latency
by $13.9\%$, $34.8\%$, and $70.1\%$, respectively.
All three directions remain faster on every measured context from
approximately 4K to 19K tokens. The benefit depends on context length:
Qwen3 $4\mathrm{B}\!\to\!8\mathrm{B}$ is slower on the shortest sample.
In all measured reverse-direction cases, source prefill alone exceeds target
prefill, so newly encoding with a larger source offers no latency benefit.
These results favor smaller-to-larger transfer for constructing target states
from scratch; transferring already available states remains useful in both
directions, as evaluated above.

\begin{table}[!ht]
\centering\footnotesize
\setlength{\tabcolsep}{3pt}
\renewcommand{\arraystretch}{1.08}
\caption{Target-state construction latency (ms), including source encoding, on the shortest and longest contexts.}
\label{tab:latency-with-source}
\begin{tabular*}{\textwidth}{@{\extracolsep{\fill}}l*{6}{r}@{}}
\toprule
Direction & $L$ & Source & Transfer & Total & Target & Speedup \\
\midrule
Qwen3 $4\mathrm{B}\!\to\!8\mathrm{B}$ & 1001 & 69.4 & 36.4 & 105.8 & 91.1 & $0.86\times$ \\
 & 19070 & 1496.7 & 219.6 & 1716.3 & 1993.4 & $1.16\times$ \\
\midrule
Llama $3\mathrm{B}\!\to\!8\mathrm{B}$ & 995 & 42.6 & 33.6 & 76.2 & 86.4 & $1.13\times$ \\
 & 19047 & 995.8 & 202.9 & 1198.7 & 1837.5 & $1.53\times$ \\
\midrule
Qwen2.5 $7\mathrm{B}\!\to\!32\mathrm{B}$ & 1001 & 85.7 & 52.3 & 138.0 & 308.5 & $2.24\times$ \\
 & 19070 & 1607.3 & 397.4 & 2004.7 & 6706.3 & $3.35\times$ \\
\midrule
Qwen3 $8\mathrm{B}\!\to\!4\mathrm{B}$ & 1001 & 102.5 & 36.5 & 139.0 & 69.3 & $0.50\times$ \\
 & 19070 & 2127.7 & 222.2 & 2349.9 & 1456.3 & $0.62\times$ \\
\midrule
Llama $8\mathrm{B}\!\to\!3\mathrm{B}$ & 995 & 90.7 & 35.4 & 126.1 & 39.8 & $0.32\times$ \\
 & 19047 & 1905.4 & 173.9 & 2079.3 & 937.4 & $0.45\times$ \\
\midrule
Qwen2.5 $32\mathrm{B}\!\to\!7\mathrm{B}$ & 1001 & 355.4 & 28.8 & 384.2 & 76.6 & $0.20\times$ \\
 & 19070 & 6732.5 & 114.7 & 6847.2 & 1580.7 & $0.23\times$ \\
\bottomrule
\end{tabular*}
\par\smallskip
\noindent\footnotesize Transfer includes Cartographer, Allocator, Compiler, packing, and positional rotation. Total adds source prefill; speedup is Target/Total. Totals are computed from the reported stage medians before final display rounding.
\end{table}
\FloatBarrier

\end{document}